\documentclass{sig-alternate-05-2015}
\pdfoutput=1
\usepackage{CJKutf8}

\usepackage{epstopdf}
\usepackage{multirow}
\usepackage{balance}
\usepackage{amsfonts}
\usepackage{amsmath}
\usepackage{amssymb}
\usepackage{graphicx}
\usepackage{subfigure}
\usepackage[noend,ruled]{algorithm2e}
\usepackage{microtype}
\usepackage{lmodern}
\usepackage{url}
\usepackage{enumitem}
\usepackage{wrapfig}
\usepackage[table]{xcolor}

\newtheorem{thm:def}{Definition}
\newtheorem{thm:eg}{Example}
\newtheorem{thm:lem}{Lemma}
\newtheorem{thm:obs}{Observation}

\newcommand{\nop}[1]{}

\newcommand{\mquote}[1]{{``\emph{#1}''}}
\newcommand{\SegPhrase}{\mbox{\sf SegPhrase}\xspace}

\newcommand{\LIPMine}{\mbox{\sf AutoPhrase}\xspace}
\newcommand{\ie}{{\sl i.e.}}
\newcommand{\eg}{{\sl e.g.}}

\newcommand{\etal}{{\sl et al.}}
\newcommand{\trans}{{\delta}}
\newcommand{\phrase}[1]{{$\lceil$#1$\rfloor$}}
\newcommand{\term}[1]{\textbf{\emph{#1}}}
\newcommand{\method}[1]{\mbox{\sf #1}\xspace}
\newcommand{\MHE}{minimal human effort\xspace}

\DeclareMathAlphabet{\mathbbold}{U}{bbold}{m}{n}

\begin{document}

\title{Automated Phrase Mining from Massive Text Corpora}

\numberofauthors{1}
\author{
\alignauthor
Jingbo Shang$^1$, Jialu Liu$^2$, Meng Jiang$^1$, Xiang Ren$^1$, Clare R Voss$^3$, Jiawei Han$^1$
\and
\alignauthor
\affaddr{$^1$Computer Science Department, University of Illinois at Urbana-Champaign, IL, USA}
\and
\alignauthor
\affaddr{$^2$Google Research, New York City, NY, USA}
\and
\alignauthor
\affaddr{$^3$Computational \& Information Sciences Directorate, Army Research Laboratory}
\and
\alignauthor
\email{$^1$\{shang7, mjiang89, xren7, hanj\}@illinois.edu $\quad$ $^2$jialu@google.com $\quad$ $^3$clare.r.voss.civ@mail.mil}
}

\clubpenalty=10000
\widowpenalty=10000
\predisplaypenalty=10000

\maketitle

\begin{abstract}

As one of the fundamental tasks in text analysis, phrase mining aims at extracting quality phrases from a text corpus. Phrase mining is important in various tasks such as information extraction/retrieval, taxonomy construction, and topic modeling.
Most existing methods rely on complex, trained linguistic analyzers, and thus likely have unsatisfactory performance on text corpora of new domains and genres without extra but expensive adaption.
Recently, a few data-driven methods have been developed successfully for extraction of phrases from massive domain-specific text. However, none of the state-of-the-art models is fully automated because they require human experts for designing rules or labeling phrases. 

Since one can easily obtain many quality phrases from public knowledge bases to a scale that is much larger than that produced by human experts, in this paper, we propose a novel framework for automated phrase mining, \LIPMine, which leverages this large amount of high-quality phrases in an effective way and achieves better performance compared to limited human labeled phrases.
In addition, we develop a POS-guided phrasal segmentation model, which incorporates the shallow syntactic information in part-of-speech (POS) tags to further enhance the performance, when a POS tagger is available.
Note that, \LIPMine can support any language as long as a general knowledge base (\eg, Wikipedia) in that language is available, while benefiting from, but not requiring, a POS tagger.
Compared to the state-of-the-art methods, the new method has shown significant improvements in effectiveness on five real-world datasets across different domains and languages.

\end{abstract}

\section{Introduction}

Phrase mining refers to the process of automatic extraction of high-quality phrases (\eg, scientific terms and general entity names) in a given corpus (\eg, research papers and news). Representing the text with \emph{quality phrases} instead of \emph{$n$-grams} can improve computational models for applications such as information extraction/retrieval, taxonomy construction, and topic modeling.

Almost all the state-of-the-art methods, however, require human experts at certain levels. Most existing methods~\cite{frantzi2000automatic,park2002automatic,zhang2008comparative} rely on \emph{complex, trained linguistic analyzers} (\eg, dependency parsers) to locate phrase mentions, and thus may have unsatisfactory performance on text corpora of new domains and genres without extra but expensive adaption. Our latest domain-independent method SegPhrase~\cite{sigmod15_liu} outperforms many other approaches~\cite{frantzi2000automatic,park2002automatic,zhang2008comparative,deane2005nonparametric,Aditya10,ahmedTopMine2015}, but still needs \emph{domain experts} to first carefully select hundreds of varying-quality phrases
from millions of candidates, and then annotate them with binary labels. 

Such reliance on manual efforts by domain and linguistic experts becomes an impediment for timely analysis of massive, emerging text corpora in specific domains. 
An ideal \emph{automated phrase mining} method is supposed to be \emph{domain-independent}, \emph{with \MHE or reliance on linguistic analyzers}\footnote{The phrase ``\MHE'' indicates using only existing general knowledge bases without any other human effort.}. 
Bearing this in mind, we propose a novel automated phrase mining framework \LIPMine in this paper, going beyond SegPhrase, to further get rid of additional manual labeling effort and enhance the performance, mainly using the following two new techniques.
\begin{enumerate}[leftmargin=*,noitemsep,nolistsep]
\item \emph{Robust Positive-Only Distant Training.}
In fact, many high-quality phrases are freely available in general knowledge bases, and they can be easily obtained to a scale that is much larger than that produced by human experts. 
Domain-specific corpora usually contain some quality phrases also encoded in general knowledge bases, even when there may be no other domain-specific knowledge bases.
Therefore, for distant training, we leverage the existing high-quality phrases, as available from general knowledge bases, such as Wikipedia and Freebase, to get rid of additional manual labeling effort.
We independently build samples of positive labels from general knowledge bases and negative labels from the given domain corpora, and train a number of base classifiers. 
We then aggregate the predictions from these classifiers, whose independence helps reduce the noise from negative labels.
\item \emph{POS-Guided Phrasal Segmentation.}
There is a trade-off between the performance and domain-independence when incorporating linguistic processors in the phrase mining method.
On the domain independence side, the accuracy might be limited without linguistic knowledge. It is difficult to support multiple languages, if the method is completely language-blind.
On the accuracy side, relying on complex, trained linguistic analyzers may hurt the domain-independence of the phrase mining method. For example, it is expensive to adapt dependency parsers to special domains like clinical reports.
As a compromise, we propose to incorporate a \emph{pre-trained} part-of-speech (POS) tagger to further enhance the performance, when it is available for the language of the document collection.
The POS-guided phrasal segmentation leverages the shallow syntactic information in POS tags to guide the phrasal segmentation model locating the boundaries of phrases more accurately.
\end{enumerate}

In principle, \LIPMine can support any language as long as a general knowledge base in that language is available.
In fact, at least 58 languages have more than 100,000 articles in Wikipedia as of Feb, 2017\footnote{\url{https://meta.wikimedia.org/wiki/List_of_Wikipedias}}.
Moreover, since pre-trained part-of-speech (POS) taggers are widely available in many languages (\eg, more than 20 languages in TreeTagger~\cite{schmid1995treetagger}\footnote{\url{http://www.cis.uni-muenchen.de/~schmid/tools/TreeTagger/}}), the POS-guided phrasal segmentation can be applied in many scenarios.
It is worth mentioning that for domain-specific knowledge bases (\eg, MeSH terms in the biomedical domain) and trained POS taggers, the same paradigm applies.
In this study, without loss of generality, we only assume the availability of a general knowledge base together with a pre-trained POS tagger.

As demonstrated in our experiments, \LIPMine not only works effectively in multiple domains like scientific papers, business reviews, and Wikipedia articles, but also supports multiple languages, such as English, Spanish, and Chinese. 

Our main contributions are highlighted as follows:
\begin{itemize}[leftmargin=*,noitemsep,nolistsep]
\item We study an important problem, \emph{automated phrase mining}, and analyze its major challenges as above.
\item We propose a robust positive-only distant training method for phrase quality estimation to minimize the human effort.
\item We develop a novel phrasal segmentation model to leverage POS tags to achieve further improvement, when a POS tagger is available.
\item We demonstrate the robustness and accuracy of our method and show improvements over prior methods, with results of experiments conducted on five real-world datasets in different domains (scientific papers, business reviews, and Wikipedia articles) and different languages (English, Spanish, and Chinese).
\end{itemize}
The rest of the paper is organized as follows. Section~\ref{sec:rel} positions our work relative to existing works. Section~\ref{sec:pre} defines basic concepts including four requirements of phrases. The details of our method are covered in Section~\ref{sec:method}. Extensive experiments and case studies are presented in Section~\ref{sec:exp}.
We conclude the study in Section~\ref{sec:con}.


\begin{figure*}[t!]
  \centering
  \includegraphics[width=0.95\textwidth]{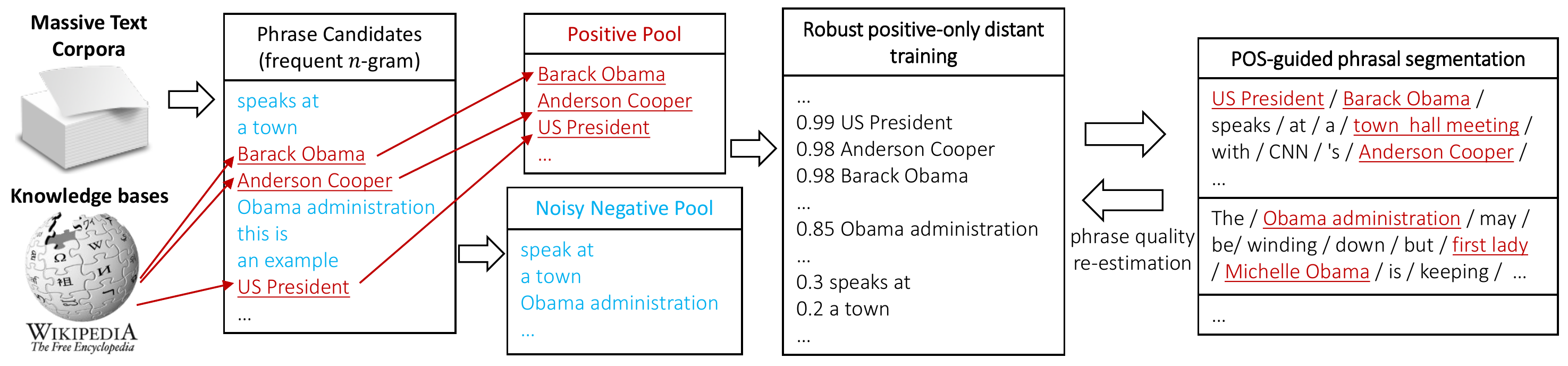}
  \vspace{-0.3cm}
  \caption{The overview of \LIPMine. The two novel techniques developed in this paper are highlighted.}
  \label{fig:overview}
  \vspace{-0.5cm}
\end{figure*}

\section{Related Work}\label{sec:rel}

Identifying quality phrases efficiently has become ever more central and critical for effective handling of massively increasing-size text datasets. 
In contrast to keyphrase extraction~\cite{mihalcea2004textrank,witten1999kea,liu2011automatic}, this task goes beyond the scope of single documents and provides useful cross-document signals.
The natural language processing~(NLP) community has conducted extensive studies typically referred to as automatic term recognition~\cite{frantzi2000automatic,park2002automatic,zhang2008comparative}, for the computational task of extracting terms (such as technical phrases).
This topic also attracts attention in the information retrieval~(IR) community~\cite{evans1996noun,Aditya10} since selecting appropriate indexing terms is critical to the improvement of search engines where the ideal indexing units represent the main concepts in a corpus, not just literal bag-of-words.

Text indexing algorithms typically filter out stop words and restrict candidate terms to noun phrases.
With pre-defined part-of-speech~(POS) rules, one can identify noun phrases as term candidates in POS-tagged documents.
Supervised noun phrase chunking techniques~\cite{punyakanok2001use,xun2000unified,chen1994extracting}  exploit such tagged documents to automatically learn rules for identifying noun phrase boundaries. Other methods may utilize more sophisticated NLP technologies such as dependency parsing to further enhance the precision~\cite{koo2008simple,mcdonald2005non}. With candidate terms collected, the next step is to leverage certain statistical measures derived from the corpus to estimate phrase quality. Some methods rely on other reference corpora for the calibration of ``termhood'' \cite{zhang2008comparative}.
The dependency on these various kinds of linguistic analyzers, domain-dependent language rules, and expensive human labeling, makes it challenging to extend these approaches to emerging, big, and unrestricted corpora, which may include many different domains, topics, and languages.

To overcome this limitation, data-driven approaches opt instead to make use of frequency statistics in the corpus to address both candidate generation and quality estimation~\cite{deane2005nonparametric,Aditya10,ahmedTopMine2015,sigmod15_liu}.
They do not rely on complex linguistic feature generation, domain-specific rules or extensive labeling efforts. Instead, they rely on large corpora containing hundreds of thousands of documents to help deliver superior performance~\cite{sigmod15_liu}.
In~\cite{Aditya10}, several indicators, including frequency and comparison to super/sub-sequences, were proposed to extract $n$-grams that reliably indicate frequent, concise concepts.
Deane~\cite{deane2005nonparametric} proposed a heuristic metric over frequency distribution based on Zipfian ranks, to measure lexical association for phrase candidates.
As a preprocessing step towards topical phrase extraction, El-Kishky \etal~\cite{ahmedTopMine2015} proposed to mine \emph{significant phrases} based on frequency as well as document context following a bottom-up fashion, which only considers a part of quality phrase criteria, \emph{popularity} and \emph{concordance}.
Our previous work~\cite{sigmod15_liu} succeeded at integrating phrase quality estimation with
phrasal segmentation to further rectify the initial set of statistical features, based on local occurrence context.
Unlike previous methods which are purely unsupervised, ~\cite{sigmod15_liu} required a small set of phrase labels to train its phrase quality estimator.
It is worth mentioning that all these approaches still depend on the human effort (\eg, setting domain-sensitive thresholds). Therefore, extending them to work automatically is challenging.


\section{Preliminaries}\label{sec:pre}

The goal of this paper is to develop an automated phrase mining method to extract quality phrases from a large collection of documents without human labeling effort, and with only limited, shallow linguistic analysis.
The main input to the automated phrase mining task is a corpus and a knowledge base. The input corpus is a textual word sequence in a particular language and a specific domain, of arbitrary length. The output is a ranked list of phrases with decreasing quality.

The \LIPMine framework is shown in Figure~\ref{fig:overview}. The work flow is completely different form our previous domain-independent phrase mining method requiring human effort~\cite{sigmod15_liu}, although the phrase candidates and the features used during phrase quality (re-)estimation are the same. In this paper, we propose a robust positive-only distant training to minimize the human effort and develop a POS-guided phrasal segmentation model to improve the model performance. In this section, we briefly introduce basic concepts and components as preliminaries.

A \term{phrase} is defined as a sequence of words that appear consecutively in the text, forming a complete semantic unit in certain contexts of the given documents~\cite{finch2000linguistic}.
The \term{phrase quality} is defined to be the probability of a word sequence being a complete semantic unit, meeting the following criteria~\cite{sigmod15_liu}:
\begin{itemize}[leftmargin=*,noitemsep,nolistsep]
\item \term{Popularity}: Quality phrases should occur with sufficient frequency in the \emph{given} document collection.
\item \term{Concordance}: The collocation of tokens in quality phrases occurs with significantly higher probability than expected due to chance.
\item \term{Informativeness}: A phrase is informative if it is indicative of a specific topic or concept.
\item \term{Completeness}: Long frequent phrases and their subsequences within those phrases may both satisfy the 3 criteria above. A phrase is deemed complete when it can be interpreted as a complete semantic unit in some given document context. Note that a phrase and a subphrase contained within it, may both be deemed complete, depending on the context in which they appear. For example, ``\emph{relational database system}'', ``\emph{relational database}'' and ``\emph{database system}'' can all be valid in certain context.
\end{itemize}
\LIPMine will estimate the phrase quality based on the positive and negative pools twice, once before and once after the POS-guided phrasal segmentation. That is, the POS-guided phrasal segmentation requires an initial set of phrase quality scores; we estimate the scores based on raw frequencies beforehand; and then once the feature values have been rectified, we re-estimate the scores. 

Only the phrases satisfying all above requirements are recognized as \term{quality phrases}.
\begin{thm:eg}
``\emph{strong tea}'' is a quality phrase while ``\emph{heavy tea}'' fails to be due to \emph{concordance}.
``\emph{this paper}'' is a \emph{popular} and \emph{concordant} phrase, but is not \emph{informative} in research publication corpus.
``\emph{NP-complete in the strong sense}'' is a quality phrase while ``\emph{NP-complete in the strong}'' fails to be due to \emph{completeness}.
$\square$
\end{thm:eg}

To automatically mine these quality phrases, the first phase of \LIPMine (see leftmost box in Figure~\ref{fig:overview}) establishes the set of \term{phrase candidates} that contains all $n$-grams over the minimum support threshold $\tau$ (\eg, 30) in the corpus. Here, this threshold refers to \term{raw frequency} of the $n$-grams calculated by string matching.
In practice, one can also set a phrase length threshold (\eg, $n \le 6$) to restrict the number of words in any phrase.
Given a phrase candidate $w_1 w_2 \ldots w_n$, its phrase quality is:
\begin{equation*}
\vspace{-0.1cm}
	Q(w_1w_2\ldots w_n) = p(\lceil w_1w_2\ldots w_n\rfloor|w_1w_2\ldots w_n) \in [0,1]
\end{equation*}
where $\lceil w_1w_2\ldots w_n \rfloor$ refers to the event that these words constitute a phrase. $Q(\cdot)$, also known as the \term{phrase quality estimator}, is initially learned from data based on statistical features\footnote{See \url{https://github.com/shangjingbo1226/AutoPhrase} for further details}, such as point-wise mutual information, point-wise KL divergence, and inverse document frequency, designed to model concordance and informativeness mentioned above.
Note the phrase quality estimator is computed independent of POS tags.
For unigrams, we simply set their phrase quality as $1$.

\begin{thm:eg} A good quality estimator will return $Q(\text{this paper}) \approx 0$ and $Q(\text{relational database system}) \approx 1$. $\square$
\end{thm:eg}

Then, to address the completeness criterion, the \term{phrasal segmentation} finds the best segmentation for each sentence.

\begin{thm:eg} Ideal phrasal segmentation results are as follows.
\begin{center}
\scalebox{0.8}{
\emph{
    \begin{tabular}{l|l}
        \#1: & ... / the / Great Firewall / is / ... \\
        \hline
        \#2: & This / is / a / great / firewall software/ . \\
        \hline
        \#3: & The / discriminative classifier / SVM / is / ... \\
    \end{tabular}
}
}
\end{center}
$\square$
\end{thm:eg}

During the \term{phrase quality re-estimation}, related statistical features will be re-computed based on the \term{rectified frequency} of phrases, which means the number of times that a phrase becomes a complete semantic unit in the identified segmentation. After incorporating the rectified frequency, the phrase quality estimator $Q(\cdot)$ also models the \emph{completeness} in addition to \emph{concordance} and \emph{informativeness}.

\begin{thm:eg}
Continuing the previous example, the \emph{raw frequency} of the phrase ``\emph{great firewall}'' is $2$, but its \emph{rectified frequency} is $1$. 
Both the \emph{raw frequency} and the \emph{rectified frequency} of the phrase ``\emph{firewall software}'' are $1$. 
The \emph{raw frequency} of the phrase ``\emph{classifier SVM}'' is $1$, but its \emph{rectified frequency} is $0$.
$\square$
\end{thm:eg}


\section{Methodology}\label{sec:method}

In this section, we focus on introducing our two new techniques.

\subsection{Robust Positive-Only Distant Training} \label{sec:method:distant}

To estimate the phrase quality score for each phrase candidate, our previous work~\cite{sigmod15_liu} required domain experts to first carefully select hundreds of varying-quality phrases
from millions of candidates, and then annotate them with binary labels. 
For example, for computer science papers, our domain experts provided hundreds of positive labels (\eg, \mquote{spanning tree} and \mquote{computer science}) and negative labels (\eg, \mquote{paper focuses} and \mquote{important form of}).
However, creating such a label set is expensive, especially in specialized domains like clinical reports and business reviews, because this approach provides no clues for how to identify the phrase candidates to be labeled.
In this paper, we introduce a method that only utilizes existing general knowledge bases without any other human effort.

\subsubsection{Label Pools}

Public knowledge bases (\eg, Wikipedia) usually encode a considerable number of high-quality phrases in the titles, keywords, and internal links of pages. For example, by analyzing the internal links and synonyms\footnote{\url{https://github.com/kno10/WikipediaEntities}} in English Wikipedia, more than a hundred thousand high-quality phrases were discovered. 
As a result, we place these phrases in a \term{positive pool}.

Knowledge bases, however, rarely, if ever, identify phrases that fail to meet our criteria, what we call \emph{inferior phrases}. An important observation is that the number of phrase candidates, based on \emph{$n$-grams} (recall leftmost box of Figure~\ref{fig:overview}), is huge and the majority of them are actually of of inferior quality (\eg, ``Francisco opera and''). In practice, based on our experiments, among millions of phrase candidates, usually, only about 10\% are in good quality. Therefore, phrase candidates that are derived from the given corpus but that fail to match any high-quality phrase derived from the given knowledge base, are used to populate a large but noisy \term{negative pool}.

\subsubsection{Noise Reduction}\label{sec:noisy_training}

\begin{figure}[t]
      \centering
      \includegraphics[width=0.47\textwidth]{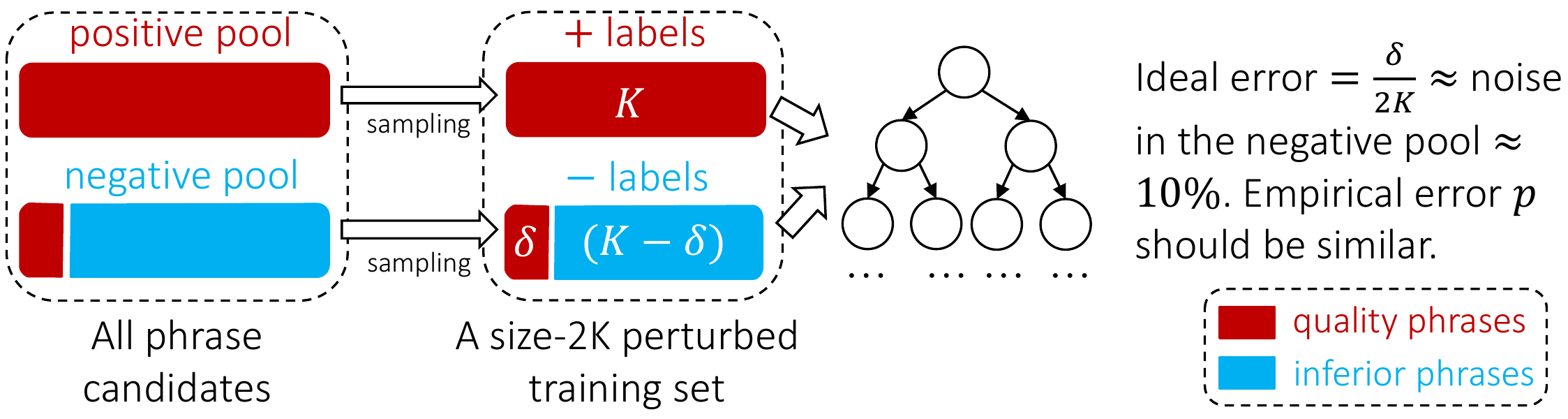}
      \vspace{-0.3cm}
      \caption{The illustration of each base classifier. In each base classifier, we first randomly sample $K$ positive and negative labels from the pools respectively. There might be $\delta$ quality phrases among the $K$ negative labels. An unpruned decision tree is trained based on this perturbed training set.}\label{fig:noisy_training}
      \vspace{-0.5cm}
\end{figure}

Directly training a classifier based on the noisy label pools is not a wise choice: some phrases of high quality from the given corpus may have been missed (\ie, inaccurately binned into the negative pool) simply because they were not present in the knowledge base. 
Instead, we propose to utilize an ensemble classifier that averages the results of $T$ independently trained base classifiers.
As shown in Figure~\ref{fig:noisy_training}, for each base classifier, we randomly draw $K$ phrase candidates with replacement from the positive pool and the negative pool respectively (considering a canonical balanced classification scenario). 
This size-$2K$ subset of the full set of all phrase candidates is called a \term{perturbed training set}~\cite{breiman2000randomizing}, because the labels of some ($\delta$ in the figure) quality phrases are switched from positive to negative.
In order for the ensemble classifier to alleviate the effect of such noise, we need to use base classifiers with the lowest possible training errors.
We grow an unpruned decision tree to the point of separating all phrases to meet this requirement. 
In fact, such decision tree will always reach $100\%$ training accuracy when no two positive and negative phrases share identical feature values in the perturbed training set. 
In this case, its ideal error is $\frac{\delta}{2K}$, which approximately equals to the proportion of switched labels among all phrase candidates (\ie, $\frac{\delta}{2K} \approx 10\%$). Therefore, the value of $K$ is not sensitive to the accuracy of the unpruned decision tree and is fixed as $100$ in our implementation.
Assuming the extracted features are distinguishable between quality and inferior phrases, the empirical error evaluated on all phrase candidates, $p$, should be relatively small as well. 

An interesting property of this sampling procedure is that the random selection of phrase candidates for building perturbed training sets creates classifiers that have statistically independent errors and similar erring probability~\cite{breiman2000randomizing,martinez2005switching}. 
Therefore, we naturally adopt random forest~\cite{geurts2006extremely}, which is verified, in the statistics literature, to be robust and efficient. 
The phrase quality score of a particular phrase is computed as the proportion of all decision trees that predict that phrase is a quality phrase. 
Suppose there are $T$ trees in the random forest, the ensemble error can be estimated as the probability of having more than half of the classifiers misclassifying a given phrase candidate as follows.
\begin{equation*}
        \mbox{ensemble\_ error}(T) = \sum_{t = \lfloor 1 + T / 2 \rfloor}^{T} {T \choose t} p^t(1-p)^{T-t}
\end{equation*}

\begin{wrapfigure}{r}{0.23\textwidth}
      \centering
      \vspace{-0.4cm}
      \includegraphics[width=0.21\textwidth]{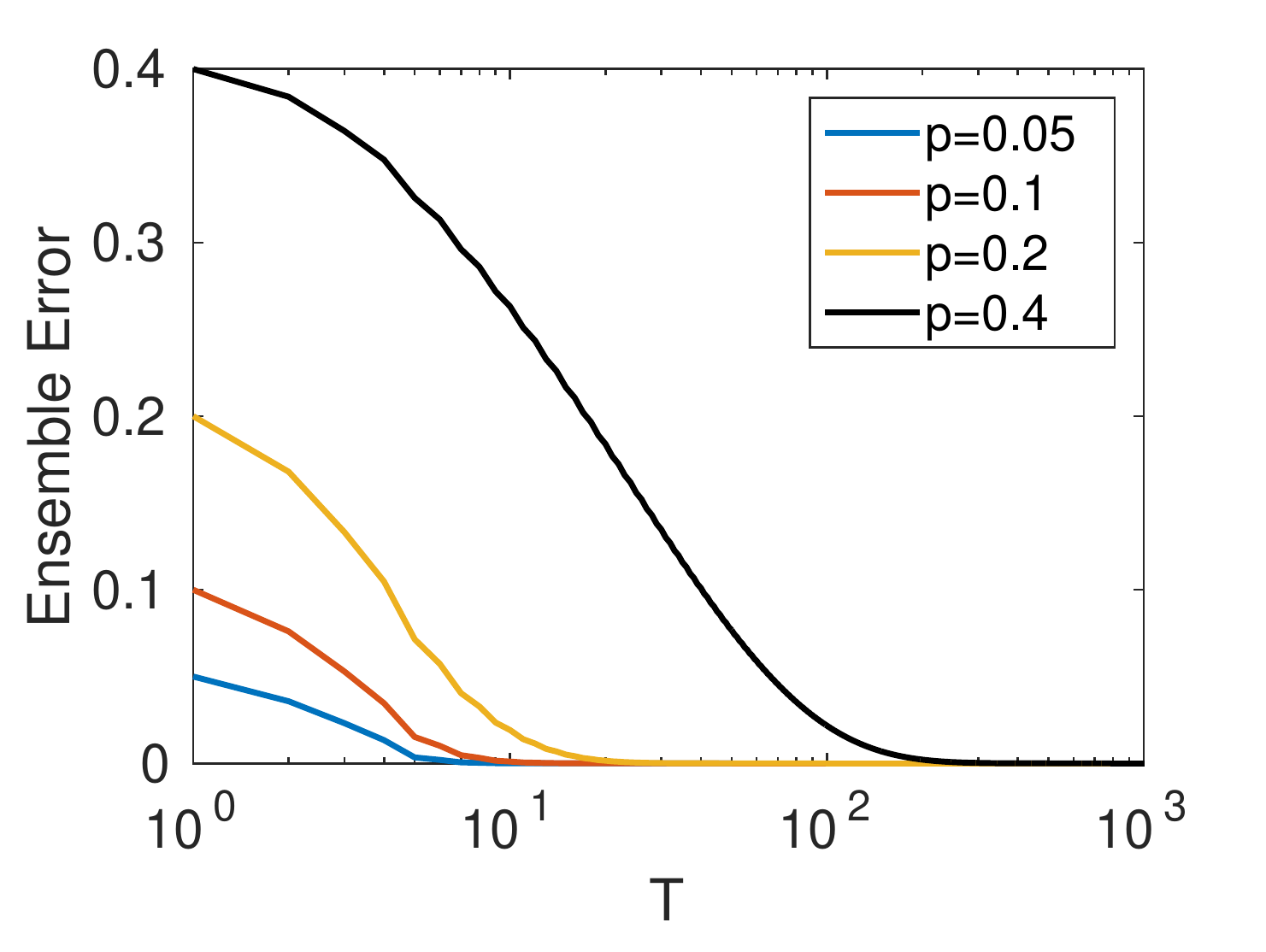}
      \vspace{-0.6cm}
      \caption{Ensemble errors of different $p$'s varying $T$.}\label{fig:ensemble}
      \vspace{-0.5cm}
\end{wrapfigure}
From Figure~\ref{fig:ensemble}, one can easily observe that the ensemble error is approaching $0$ when $T$ grows. 
In practice, $T$ needs to be set larger due to the additional error brought by model bias.
Empirical studies can be found in Figure~\ref{fig:distant_3}.

\subsection{POS-Guided Phrasal Segmentation}

\SetAlgoSkip{}
    \begin{algorithm}[t]
        \caption{\small POS-Guided Phrasal Segmentation (PGPS)}\label{alg:DP}
        \textbf{Input}: Corpus $\Omega = \Omega_1\Omega_2\ldots\Omega_n$, phrase quality $Q$, parameters $\theta_u$ and $\trans(t_x, t_y)$. \\
        \textbf{Output}: Optimal boundary index sequence $B$. \\
        \tcp{\color{blue}$h_i \equiv \max_{B} $\quad$ p( \Omega_1\Omega_2\ldots\Omega_{i-1}, B  | Q, \theta, \trans)$}
        $h_1 \leftarrow 1$, $h_i \leftarrow 0 ~(1 < i \le n + 1)$ \\
        \For{$i=1$ {\bfseries to} $n$}{
            \For{$j=i + 1$ {\bfseries to} $\min(i + \text{length threshold}, n + 1)$} {
                \tcp{\color{blue}Efficiently implemented via Trie.}
                \If {there is no phrase starting with $w_{[i, j)}$} {
                    \textbf{break} \\
                }
                \tcp{\color{blue}In practice, log and addition are used to avoid underflow.}
                \If{$h_{i} \times p(j,\lceil w_{[i, j)} \rfloor | i, t_{[i, j)}) > h_{j}$} {
                    $h_{j} \leftarrow h_{i} \times p(j,\lceil w_{[i, j)} \rfloor | i, t_{[i, j)})$ \\
                    $g_{j} \leftarrow i$
                }
            }
        }
        $j \leftarrow n + 1$, $m \leftarrow 0$ \\
        \While{$j > 1$}{
            $m \leftarrow m + 1$ \\
            $b_m \leftarrow j$ \\
            $j \leftarrow g_j$ \\
        }
        \Return $B \leftarrow 1, b_m, b_{m-1}, \ldots, b_1$
    \end{algorithm}

    Phrasal segmentation addresses the challenge of measuring \emph{completeness} (our fourth criterion) by locating all phrase mentions in the corpus and rectifying their frequencies obtained originally via string matching.

    The corpus is processed as a length-$n$ POS-tagged word sequence $\Omega = \Omega_1 \Omega_2 \ldots \Omega_n$, where $\Omega_i$ refers to a pair consisting of a word and its POS tag $\langle w_i, t_i \rangle$. A \term{POS-guided phrasal segmentation} is a partition of this sequence into $m$ segments induced by a boundary index sequence $B = \{b_1, b_2, \ldots, b_{m+1}\}$ satisfying $1$ $=$ $b_1$ $<$ $b_2$ $<$ $\dots$ $<$ $b_{m+1}$ $=$ $n$+1. The $i$-th segment refers to $\Omega_{b_i} \Omega_{b_{i}+1} \ldots \Omega_{b_{i + 1} - 1}$.

    Compared to the phrasal segmentation in our previous work~\cite{sigmod15_liu}, the POS-guided phrasal segmentation addresses the completeness requirement in a \emph{context-aware} way, instead of equivalently penalizing phrase candidates of the same length. In addition, POS tags provide shallow, language-specific knowledge, which may help boost phrase detection accuracy, especially at syntactic constituent boundaries for that language.

    Given the POS tag sequence for the full $n$-length corpus is $t = t_1 t_2 \ldots t_n$, containing the tag subsequence $t_l \ldots t_{r-1}$ (denote as $t_{[l, r)}$ for clarity), the \term{POS quality} score for that tag subsequence is defined to be the conditional probability of its corresponding word sequence being a complete semantic unit.
    Formally, we have
    \begin{equation*}
        T(t_{[l, r)}) = p(\lceil w_l \ldots w_{r} \rfloor| t) \in [0,1]
    \end{equation*}
    The POS quality score $T(\cdot)$ is designed to reward the phrases with their correctly identified POS sequences, as follows.
    \begin{thm:eg}
        Suppose the whole POS tag sequence is ``NN NN NN VB DT NN''.
        A good POS sequence quality estimator might return $T(\text{NN NN NN}) \approx 1$ and $T(\text{NN VB}) \approx 0$, where NN refers to singular or mass noun (\eg, database), VB means verb in the base form (\eg, is), and DT is for determiner (\eg, the).
        $\square$
    \end{thm:eg}

    The particular form of $T(\cdot)$ we have chosen is:
    \begin{equation*}
        T(t_{[l, r)}) = \left(1 - \trans(t_{b_r - 1}, t_{b_r})\right) \times \prod_{j=l + 1}^{r - 1} \trans( t_{j - 1}, t_{j})
    \end{equation*}
    where, $\trans(t_x, t_y)$ is the probability that the POS tag $t_x$ is adjacent and precedes POS tag $t_y$ within a phrase in the given document collection. In this formula, the first term indicates the probability that there is a phrase boundary between the words indexed $r - 1$ and $r$, while the latter product indicates the probability that all POS tags within $t_{[l, r)}$ are in the same phrase.
    This POS quality score can naturally counter the bias to longer segments because $\forall i > 1$, exactly one of $\trans(t_{i - 1}, t_{i})$ and $(1 - \trans(t_{i - 1}, t_{i}))$ is always multiplied no matter how the corpus is segmented. Note that the length penalty model in our previous work~\cite{sigmod15_liu} is a special case when all values of $\trans(t_x, t_y)$ are the same.

    Mathematically, $\trans(t_x, t_y)$ is defined as:
    \begin{equation*}
    \trans(t_x, t_y) = p(\lceil \ldots w_1 w_2  \ldots \rfloor | \Omega, \text{tag}(w_1) = t_x \wedge \text{tag}(w_2) = t_y)
    \end{equation*}
    As it depends on how documents are segmented into phrases, $\trans(t_x, t_y)$ is initialized uniformly and will be learned during the phrasal segmentation.

    Now, after we have both phrase quality $Q(\cdot)$ and POS quality $T(\cdot)$ ready, we are able to formally define the POS-guided phrasal segmentation model. The joint probability of a POS tagged sequence $\Omega$ and a boundary index sequence $B = \{b_1, b_2, \ldots, b_{m+1}\}$ is factorized as:
    \begin{equation*}
    p(\Omega, B) =  \prod_{i=1}^m p\left(b_{i+1}, \lceil w_{[b_i, b_{i+1})} \rfloor \Big|b_i, t\right)
    \end{equation*}
    where $p(b_{i+1}, \lceil w_{[b_i, b_{i+1})}\rfloor |b_i, t)$ is the probability of observing a word sequence $w_{[b_i, b_{i+1})}$ as the $i$-th quality segment given the previous boundary index $b_i$ and the whole POS tag sequence $t$.

    Since the phrase segments function as a constituent in the syntax of a sentence, they usually have weak dependence on each other~\cite{finch2000linguistic,sigmod15_liu}. As a result, we assume these segments in the word sequence are generated one by one for the sake of both efficiency and simplicity.

    For each segment, given the POS tag sequence $t$ and the start index $b_i$, the generative process is defined as follows.
    \begin{enumerate}[noitemsep,nolistsep]
        \item Generate the end index $b_{i+1}$, according to its POS quality
            \begin{equation*}
                p(b_{i+1} | b_i, t) =  T(t_{[b_i, b_{i+1})})
            \end{equation*}
        \item Given the two ends $b_i$ and $b_{i + 1}$, generate the word sequence $w_{[b_i, b_{i+1})}$ according to a multinomial distribution over all segments of length-$(b_{i+1}-b_i)$.
            \begin{equation*}
                p(w_{[b_i, b_{i+1})}|b_i, b_{i+1}) = p(w_{[b_i, b_{i+1})}| b_{i+1} - b_i)
            \end{equation*}
        \item Finally, we generate an indicator whether $w_{[b_i, b_{i+1})}$ forms a quality segment according to its quality
            \begin{equation*}
                p(\lceil w_{[b_i, b_{i+1})}\rfloor |w_{[b_i, b_{i+1})}) = Q(w_{[b_i, b_{i+1})})
            \end{equation*}
    \end{enumerate}
    We denote $p(w_{[b_i, b_{i+1})}| b_{i+1} - b_i)$ as $\theta_{w_{[b_i, b_{i+1})}}$ for convenience.
    Integrating the above three generative steps together, we have the following probabilistic factorization:
    \begin{equation*}
    \begin{split}
    & p(b_{i+1}, \lceil w_{[b_i, b_{i+1})}\rfloor | b_i, t) \\
            =& p(b_{i+1} | b_i, t) p(w_{[b_i, b_{i+1})}|b_i, b_{i+1}) p(\lceil w_{[b_i, b_{i+1})}\rfloor | w_{[b_i, b_{i+1})}) \\
            =& T(t_{[b_i, b_{i+1})}) \theta_{w_{[b_i, b_{i+1})}} Q(w_{[b_i, b_{i+1})})
    \end{split}
    \end{equation*}

    Therefore, there are three subproblems:
    \begin{enumerate}[noitemsep,nolistsep]
    \item Learn $\theta_u$ for each word and phrase candidate $u$;
    \item Learn $\trans(t_x, t_y)$ for every POS tag pair; and
    \item Infer $B$ when $\theta_{u}$ and $\trans(t_x, t_y)$ are fixed.
    \end{enumerate}

    We employ the maximum a posterior principle and maximize the joint log likelihood:
    \begin{equation}\label{eq:m-step}
    \begin{split}
    \log p(\Omega, B) &=  \sum_{i=1}^{m} \log p\left(b_{i+1}, \lceil w_{[b_i, b_{i+1})}\rfloor \Big| b_t, t\right)
    \end{split}
    \vspace{-0.1cm}
    \end{equation}

    Given $\theta_{u}$ and $\trans(t_x, t_y)$, to find the best segmentation that maximizes Equation~\eqref{eq:m-step}, we develop an efficient dynamic programming algorithm for the POS-guided phrasal segmentation as shown in Algorithm~\ref{alg:DP}.

    When the segmentation $S$ and the parameter $\theta$ are fixed, the closed-form solution of $\trans(t_x, t_y)$ is:
    \begin{equation} \label{eq:trans_t1_t2}
           \trans(t_x, t_y) = \frac{\sum_{i=1}^{m}\sum_{j=b_i}^{b_{i+1} - 2} \mathbbold{1}({t_j = t_x \wedge t_{j + 1} = t_y})}{\sum_{i=1}^{n - 1} \mathbbold{1}( {t_i = t_x \wedge t_{i + 1} = t_y})}
    \end{equation}
    where $\mathbbold{1}(\cdot)$ denotes the identity indicator. $\trans(t_x, t_y)$ is the unsegmented ratio among all $\langle t_x, t_y \rangle$ pairs in the given corpus.

    Similarly, once the segmentation $S$ and the parameter $\trans$ are fixed, the closed-form solution of $\theta_u$ can be derived as:
    \begin{equation} \label{eq:theta_u}
           \theta_u = \frac{\sum_{i=1}^{m} \mathbbold{1}({w_{[b_i, b_{i+1})} = u})}{\sum_{i=1}^{m} \mathbbold{1}({b_{i + 1} - b_i = |u|})}
    \end{equation}
    We can see that $\theta_u$ is the times that $u$ becomes a complete segment normalized by the number of the length-$|u|$ segments.

    \SetAlgoSkip{}
    \begin{algorithm}[t]
        \caption{Viterbi Training (VT)}\label{alg:VT}
        \textbf{Input}: Corpus $\Omega$ and phrase quality $Q$. \\
        \textbf{Output}: $\theta_u$ and $\trans(t_x, t_y)$. \\
        $\mbox{initialize $\theta$ with normalized raw frequencies in the corpus}$\\
        \While{$\theta_u$ does not converge}{
            \While{$\trans(t_x, t_y)$ does not converge}{
                $B \leftarrow$ best segmentation via Alg.~\ref{alg:DP} \\
                update $\trans(t_x, t_y)$ using $B$ according to Eq.~\eqref{eq:trans_t1_t2}\\
            }
            $B \leftarrow$ best segmentation via Alg.~\ref{alg:DP} \\
            update $\theta_u$ using $B$ according to  Eq.~\eqref{eq:theta_u}\\
        }
        \Return $\theta_u$ and $\trans(t_x, t_y)$
    \end{algorithm}

    As shown in Algorithm~\ref{alg:VT}, we choose Viterbi Training, or Hard EM in literature~\cite{allahverdyan2011comparative} to iteratively optimize parameters, because Viterbi Training converges fast and results in sparse and simple models for Hidden Markov Model-like tasks~\cite{allahverdyan2011comparative}.

    \subsection{Complexity Analysis}

    The time complexity of the most time consuming components in our framework, such as frequent $n$-gram, feature extraction, POS-guided phrasal segmentation, are all $O(|\Omega|)$ with the assumption that the maximum number of words in a phrase is a small constant (\eg, $n \le 6$), where $|\Omega|$ is the total number of words in the corpus.
    Therefore, \LIPMine is linear to the corpus size and thus being very efficient and scalable. Meanwhile, every component can be parallelized in an almost lock-free way grouping by either phrases or sentences.


\section{Experiments}\label{sec:exp}
In this section, we will apply the proposed method to mine quality phrases from five massive text corpora across three domains (scientific papers, business reviews, and Wikipedia articles) and in three languages (English, Spanish, and Chinese).
We compare the proposed method with many other methods to demonstrate its high performance.
Then we explore the robustness of the proposed positive-only distant training and its performance against expert labeling.
The advantage of incorporating POS tags in phrasal segmentation will also be proved.
In the end, we present case studies.

\subsection{Datasets}

To validate that the proposed positive-only distant training can effectively work in different domains and the POS-guided phrasal segmentation can support multiple languages effectively, we have five large collections of text in different domains and languages, as shown in Table~\ref{tbl:dataset}: Abstracts of English computer science papers from \textbf{DBLP}\footnote{\url{https://aminer.org/citation}}, English business reviews from \textbf{Yelp}\footnote{\url{https://www.yelp.com/dataset_challenge}}, Wikipedia articles\footnote{\url{https://dumps.wikimedia.org/}} in English (\textbf{EN}), Spanish (\textbf{ES}), and Chinese (\textbf{CN}).
From the existing general knowledge base Wikipedia, we extract popular mentions of entities by analyzing intra-Wiki citations within Wiki content\footnote{\url{https://github.com/kno10/WikipediaEntities}}. On each dataset, the intersection between the extracted popular mentions and the generated phrase candidates forms the positive pool. Therefore, the size of positive pool may vary in different datasets of the same language.

\begin{table}[t]
\center
\caption{Five real-world massive text corpora in different domains and multiple languages. $|\Omega|$ is the total number of words. $size_p$ is the size of positive pool.}
\label{tbl:dataset}
\scalebox{0.8}{
    \begin{tabular}{|c|c|c|c|c|c|}
    \hline
    Dataset & Domain & Language & $|\Omega|$ & File size & $size_p$ \\
    \hline
    DBLP & Scientific Paper & English & 91.6M & 618MB & 29K\\
    \hline
    Yelp & Business Review & English & 145.1M & 749MB & 22K\\
    \hline
    EN & Wikipedia Article & English & 808.0M & 3.94GB & 184K \\
    \hline
    ES & Wikipedia Article & Spanish & 791.2M & 4.06GB & 65K\\
    \hline
    CN & Wikipedia Article & Chinese & 371.9M & 1.56GB & 29K \\
    \hline
    \end{tabular}
}
\end{table}

\begin{figure*}[t!]
  \centering
  \subfigure[DBLP]{
    \includegraphics[width= 0.18\textwidth]{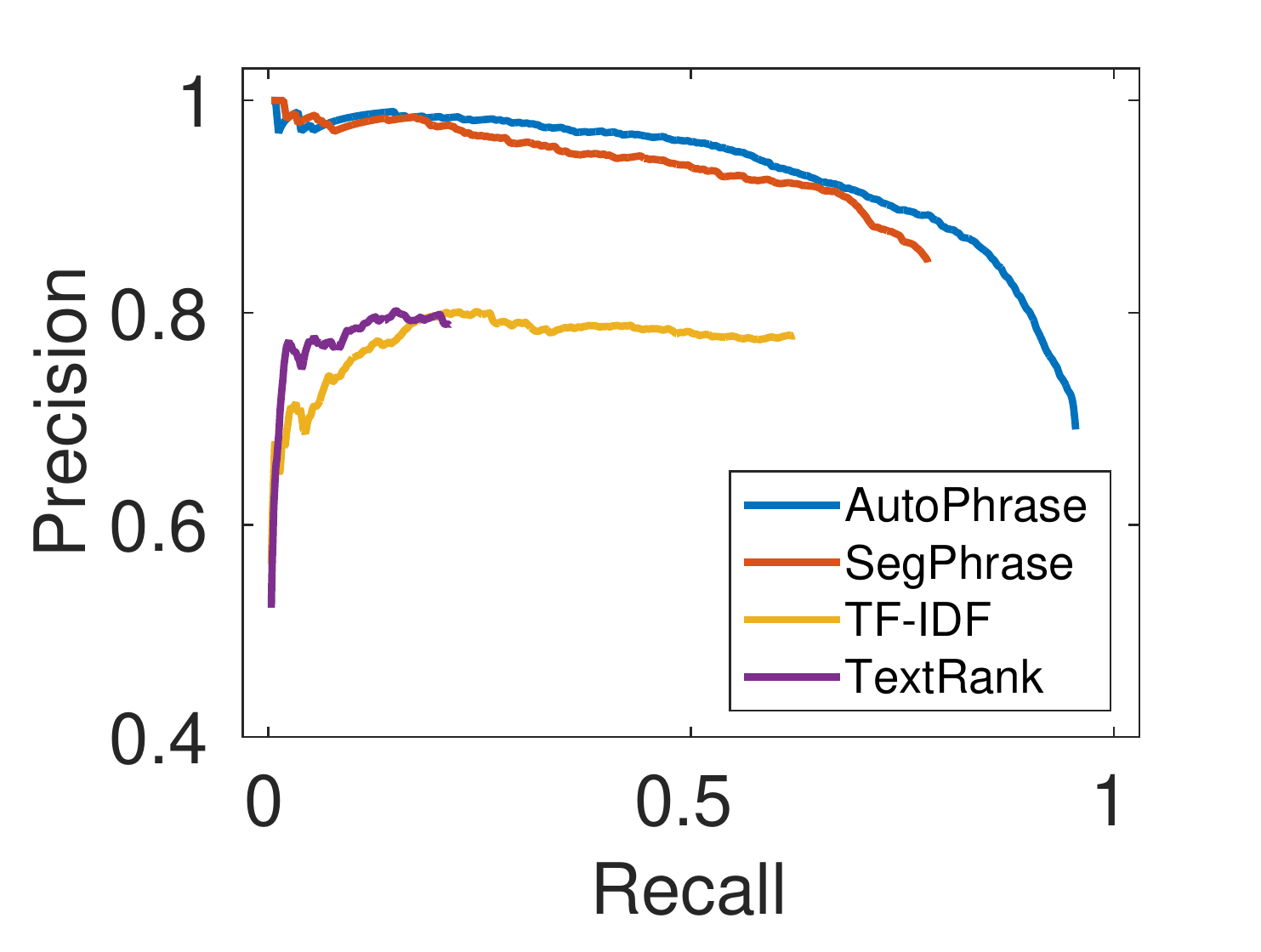}
  }
    \subfigure[Yelp]{
    \includegraphics[width= 0.18\textwidth]{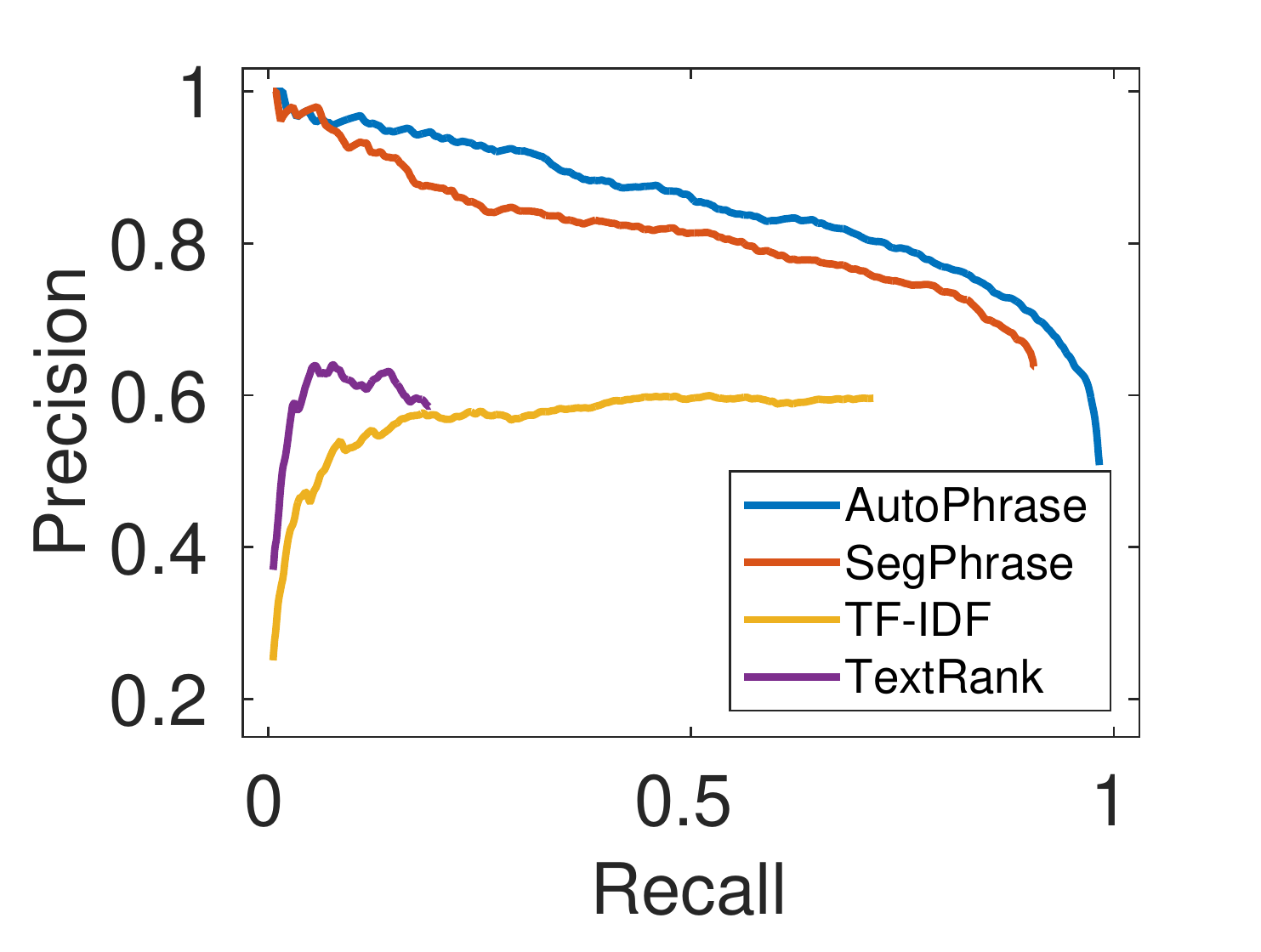}
  }
  \subfigure[EN]{
    \includegraphics[width= 0.18\textwidth]{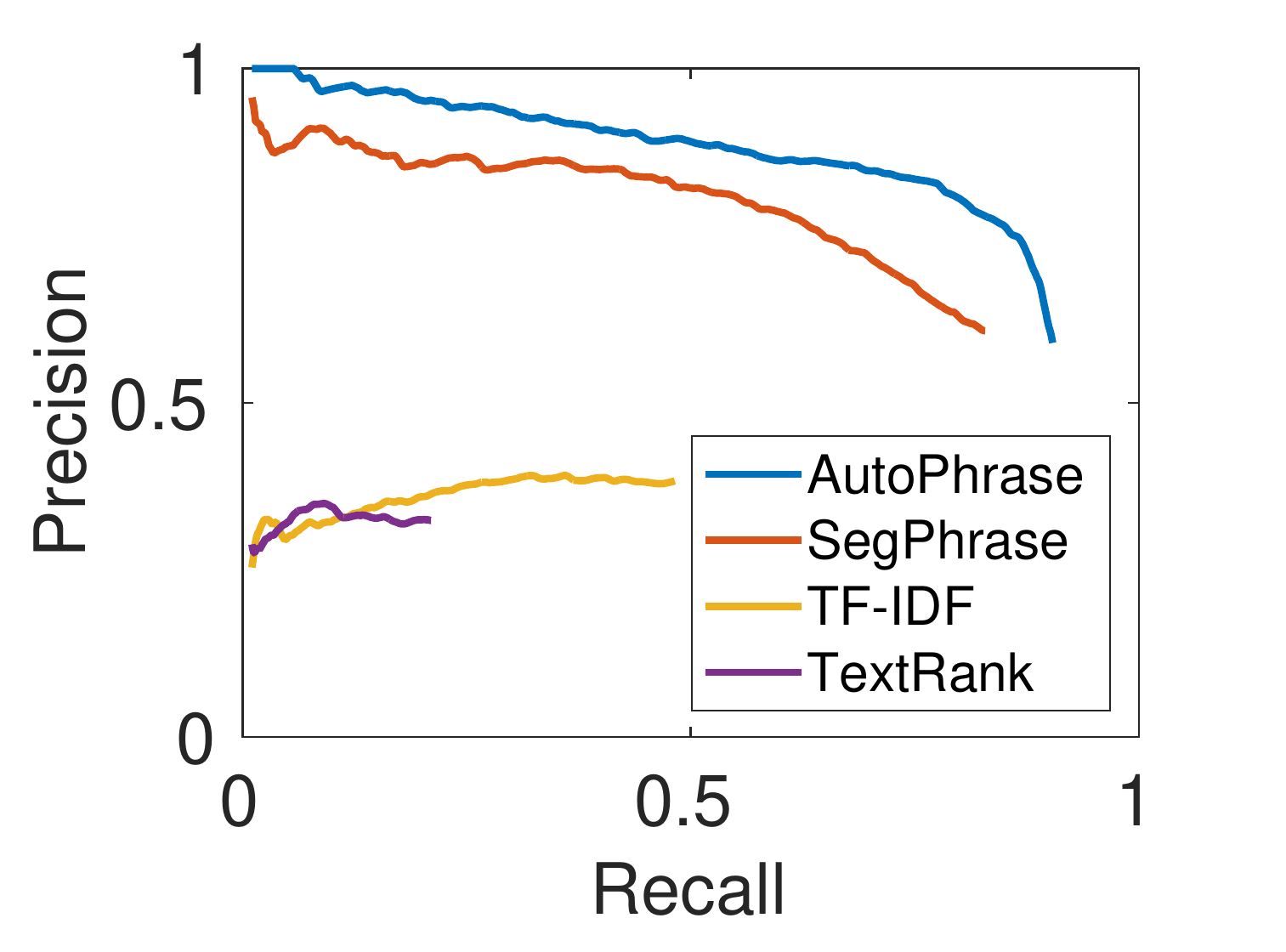}
  }
  \subfigure[ES]{
    \includegraphics[width= 0.18\textwidth]{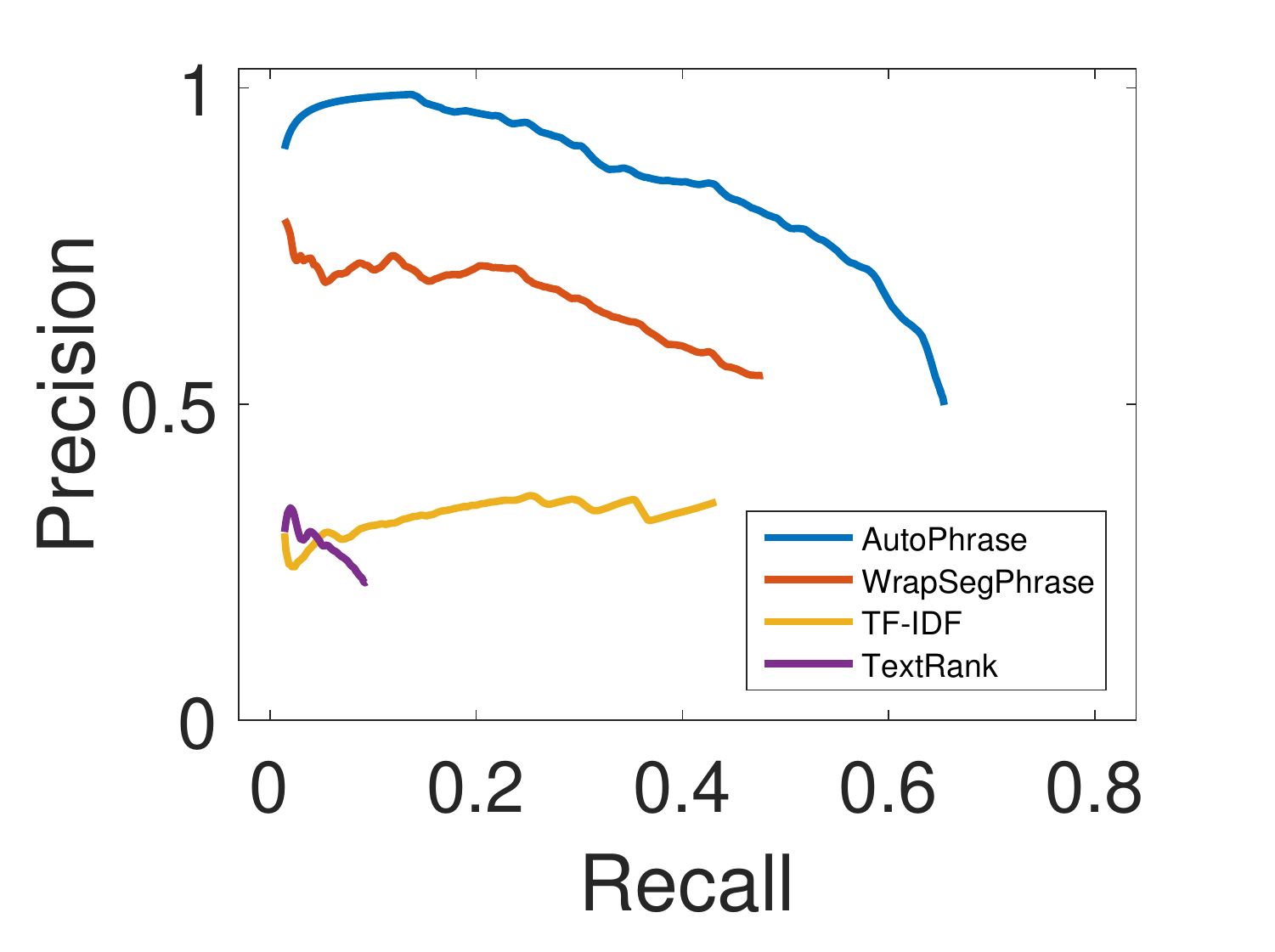}
  }
  \subfigure[CN]{
    \includegraphics[width= 0.18\textwidth]{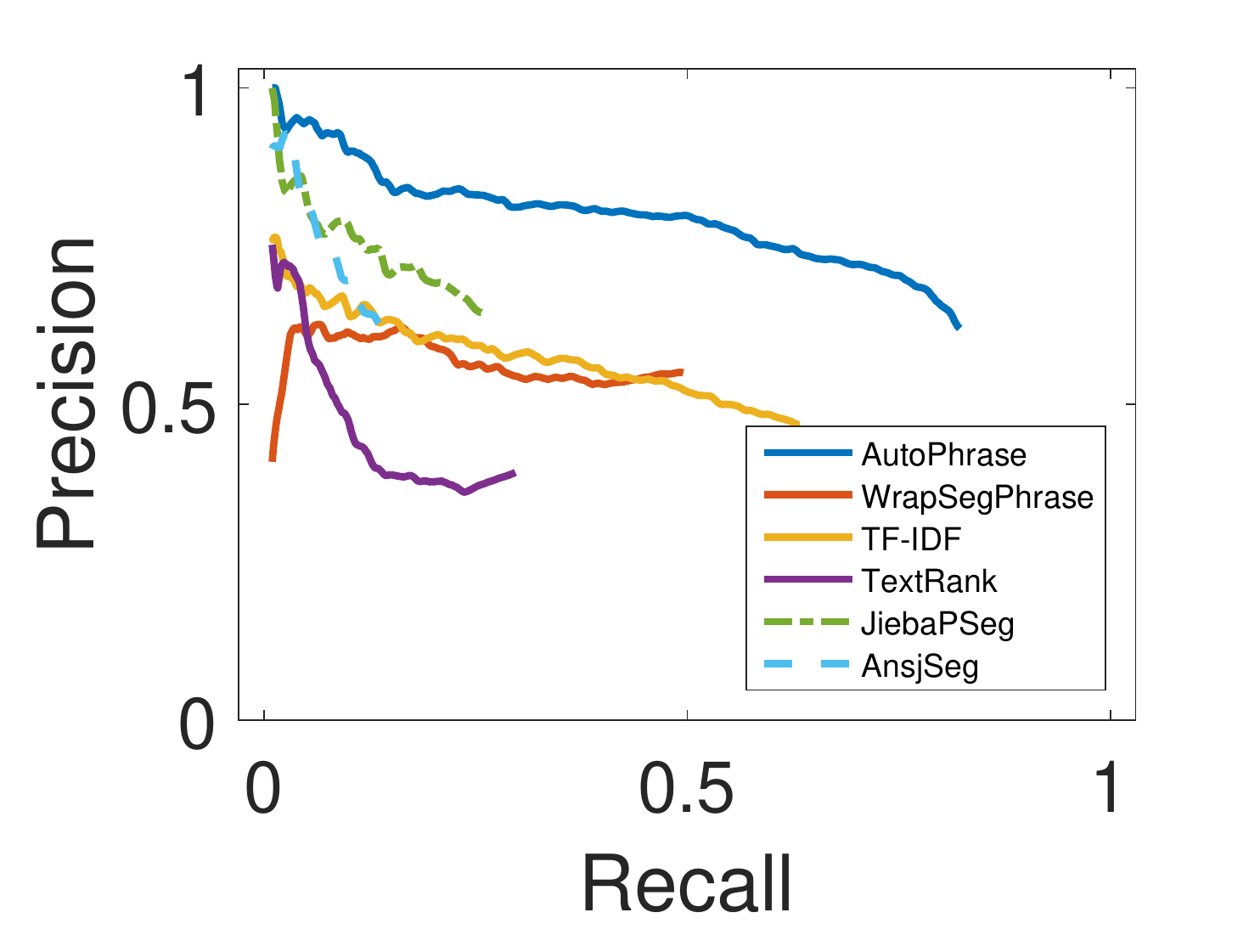}
  }
  \caption{Overall Performance Evaluation: Precision-recall curves of all methods evaluated by human annotation.}
  \label{fig:human_eval}
\end{figure*}

\subsection{Compared Methods}

We compare \textbf{AutoPhrase} with three lines of methods as follows. Every method returns a ranked list of phrases.

\noindent\textbf{SegPhrase\footnote{\url{https://github.com/shangjingbo1226/SegPhrase}}/WrapSegPhrae\footnote{\url{https://github.com/remenberl/SegPhrase-MultiLingual}}}: In English domain-specific text corpora, our latest work \SegPhrase outperformed phrase mining~\cite{ahmedTopMine2015}, keyphrase extraction~\cite{witten1999kea,Aditya10}, and noun phrase chunking methods.
\method{WrapSegPhrase} extends \SegPhrase to different languages by adding an encoding preprocessing to first transform non-English corpus using English characters and punctuation as well as a decoding postprocessing to later translate them back to the original language.
Both methods require domain expert labors.
For each dataset, we ask domain experts to annotate a representative set of 300 quality/interior phrases.

\noindent\textbf{Parser-based Phrase Extraction}: Using complicated linguistic processors, such as parsers, we can extract minimum phrase units (\eg, NP) from the parsing trees as phrase candidates. Parsers of all three languages are available in Stanford NLP tools~\cite{nivre2016universal,de2006generating,levy2003harder}. Two ranking heuristics are considered:
\begin{itemize}[leftmargin=*,noitemsep,nolistsep]
\item \textbf{TF-IDF} ranks the extracted phrases by their term frequency and inverse document frequency in the given documents;
\item \textbf{TextRank}: An unsupervised graph-based ranking model for keyword extraction~\cite{mihalcea2004textrank}.
\end{itemize}

\noindent\textbf{Pre-trained Chinese Segmentation Models}: Different from English and Spanish, phrasal segmentation in Chinese has been intensively studied because there is no space between Chinese words. The most effective and popular segmentation methods are:
\begin{itemize}[leftmargin=*,noitemsep,nolistsep]
\item \textbf{AnsjSeg}\footnote{\url{https://github.com/NLPchina/ansj_seg}} is a popular text segmentation algorithm for Chinese corpus. It ensembles statistical modeling methods of Conditional Random Fields (CRF) and Hidden Markov Models (HMMs) based on the $n$-gram setting;
\item \textbf{JiebaPSeg}\footnote{\url{https://github.com/fxsjy/jieba}} is a \emph{Chinese} text segmentation method implemented in Python. It builds a directed acyclic graph for all possible phrase combinations based on a prefix dictionary structure to achieve efficient phrase graph scanning. Then it uses dynamic programming to find the most probable combination based on the phrase frequency. For unknown phrases, an HMM-based model is used with the Viterbi algorithm.
\end{itemize}
Note that all parser-based phrase extraction and Chinese segmentation models are pre-trained based on general corpus. 

To study the effectiveness of the POS-guided segmentation, \textbf{AutoSegPhrase} adopts the length penalty instead of $\trans(t_x, t_y)$, while other components are the same as \LIPMine. \method{AutoSegPhrase} is useful when there is no POS tagger.

\subsection{Experimental Settings}

\noindent\textbf{Implementation.} The preprocessing includes tokenizers from Lucene and Stanford NLP as well as the POS tagger from TreeTagger. Our documented code package has been released and maintained in GitHub\footnote{\url{https://github.com/shangjingbo1226/AutoPhrase}}.

\noindent\textbf{Default Parameters.} We set the minimum support threshold $\sigma$ as 30. The maximum number of words in a phrase is set as $6$ according to labels from domain experts. These are two parameters required by all methods. Other parameters required by compared methods were set according to the open-source tools or the original papers.

\noindent\textbf{Human Annotation.}
We rely on human evaluators to judge the quality of the phrases which cannot be identified through any knowledge base. More specifically, on each dataset, we randomly sample $500$ such phrases from the predicted phrases of each method in the experiments. These selected phrases are shuffled in a shared \term{pool} and evaluated by 3 reviewers independently. We allow reviewers to use search engines when unfamiliar phrases encountered. By the rule of majority voting, phrases in this pool received at least two positive annotations are \emph{quality phrases}. The intra-class correlations (ICCs) are all more than 0.9 on all five datasets, which shows the agreement.

\noindent\textbf{Evaluation Metrics.} For a list of phrases, \emph{precision} is defined as the number of true quality phrases divided by the number of predicted quality phrases; \emph{recall} is defined as the number of true quality phrases divided by the total number of quality phrases. We retrieve the ranked list of the pool from the outcome of each method. When a new true quality phrase encountered, we evaluate the precision and recall of this ranked list. In the end, we plot the \emph{precision-recall curves}. In addition, \emph{area under the curve (AUC)} is adopted as another quantitative measure. AUC in this paper refers to the area under the precision-recall curve.

\subsection{Overall Performance}

Figures~\ref{fig:human_eval} presents the precision-recall curves of all compared methods evaluated by human annotation on five datasets. Overall, \LIPMine performs the best, in terms of both precision and recall.
Significant advantages can be observed, especially on two non-English datasets \emph{ES} and \emph{CN}. For example, on the \emph{ES} dataset, the recall of \LIPMine is about $20\%$ higher than the second best method (\SegPhrase) in absolute value.
Meanwhile, there is a visible precision gap between \LIPMine and the best baseline.
The phrase chunking-based methods \method{TF-IDF} and \method{TextRank} work poorly, because the extraction and ranking are modeled separately and the pre-trained complex linguistic analyzers fail to extend to domain-specific corpora. \method{TextRank} usually starts with a higher precision than \method{TF-IDF}, but its recall is very low because of the sparsity of the constructed co-occurrence graph. \method{TF-IDF} achieves a reasonable recall but unsatisfactory precision.
On the \emph{CN} dataset, the pre-trained Chinese segmentation models, \method{JiebaSeg} and \method{AnsjSeg}, are very competitive, because they not only leverage training data for segmentations, but also exhaust the engineering work, including a huge dictionary for popular Chinese entity names and specific rules for certain types of entities.
As a consequence, they can easily extract tons of well-known terms and people/location names.
Outperforming such a strong baseline further confirms the effectiveness of \LIPMine.

The comparison among the English datasets across three domains (\ie, scientific papers, business reviews, and Wikipedia articles) demonstrates that \LIPMine is reasonably \emph{domain-independent}.
The performance of parser-based methods \method{TF-IDF} and \method{TextRank} depends on the rigorous degree of the documents. For example, it works well on the \emph{DBLP} dataset but poorly on the \emph{Yelp} dataset.
However, without any human effort, \LIPMine can work effectively on domain-specific datasets, and even outperforms \SegPhrase, which is supervised by the domain experts.

The comparison among the Wikipedia article datasets in three languages (\ie, \emph{EN}, \emph{ES}, and \emph{CN}) shows that, first of all, \LIPMine \emph{supports multiple languages}. Secondly, the advantage of \LIPMine over \SegPhrase/\method{WrapSegPhrase} is more obvious on two non-English datasets \emph{ES} and \emph{CN} than the \emph{EN} dataset, which proves that \emph{the helpfulness of introducing the POS tagger}.

As conclusions, \LIPMine is able to support different domains and support multiple languages with \MHE.

\subsection{Distant Training Exploration}

To compare the distant training and domain expert labeling, we experiment with the domain-specific datasets \emph{DBLP} and \emph{Yelp}.
To be fair, all the configurations in the classifiers are the same except for the label selection process. More specifically, we come up with four training pools:
\begin{enumerate}[leftmargin=*,noitemsep,nolistsep]
\item \textbf{EP} means that domain experts give the positive pool.
\item \textbf{DP} means that a sampled subset from existing general knowledge forms the positive pool.
\item \textbf{EN} means that domain experts give the negative pool.
\item \textbf{DN} means that all \emph{unlabeled} (\ie, not in the positive pool) phrase candidates form the negative pool.
\end{enumerate}
By combining any pair of the positive and negative pools, we have four variants, \textbf{EPEN} (in \SegPhrase), \textbf{DPDN} (in \LIPMine), \textbf{EPDN}, and \textbf{DPEN}.

    \begin{figure}[t!]
      \centering
      \subfigure[DBLP]{
          \includegraphics[width= 0.22\textwidth]{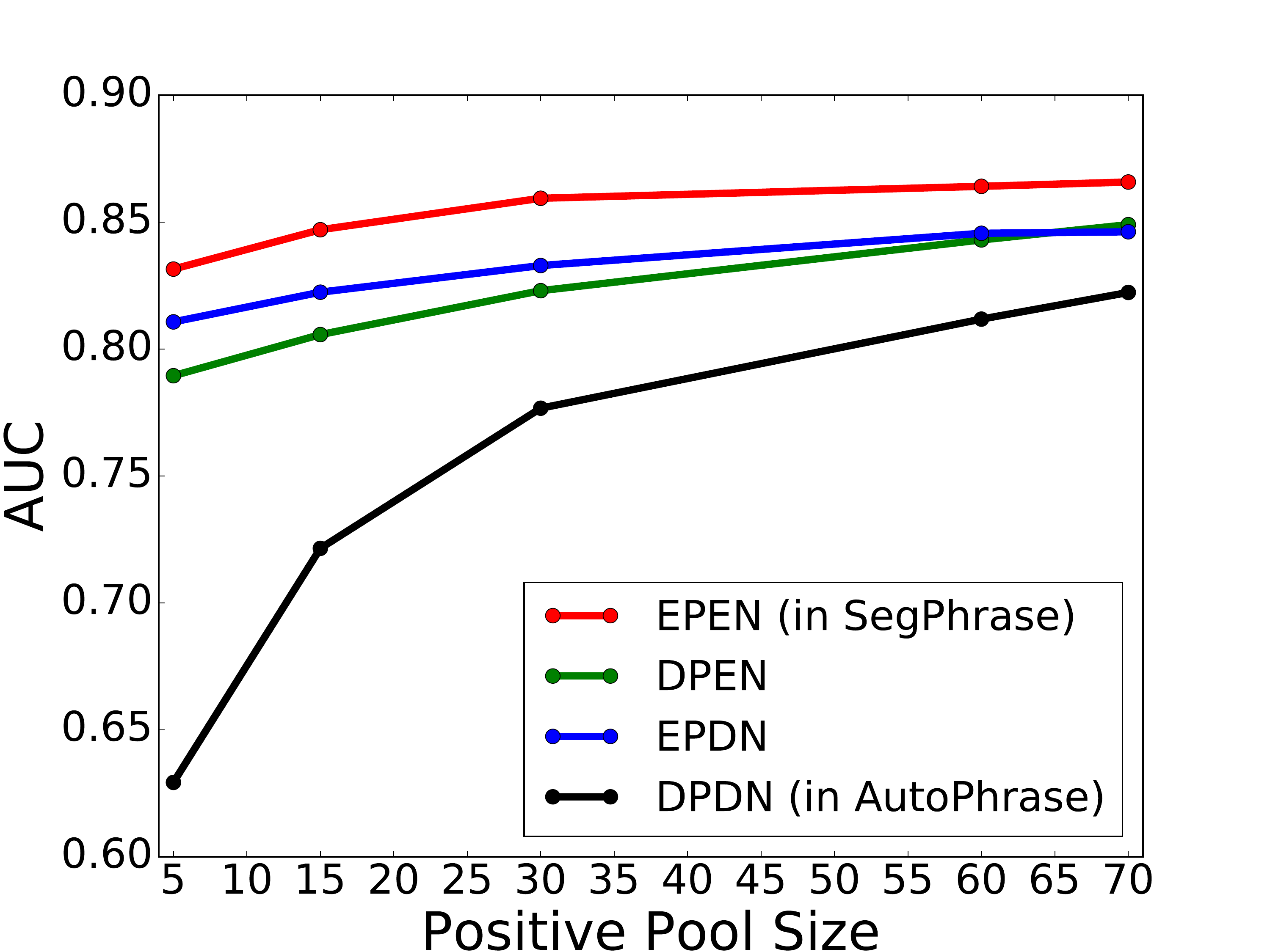}
      }
      \subfigure[Yelp]{
          \includegraphics[width= 0.22\textwidth]{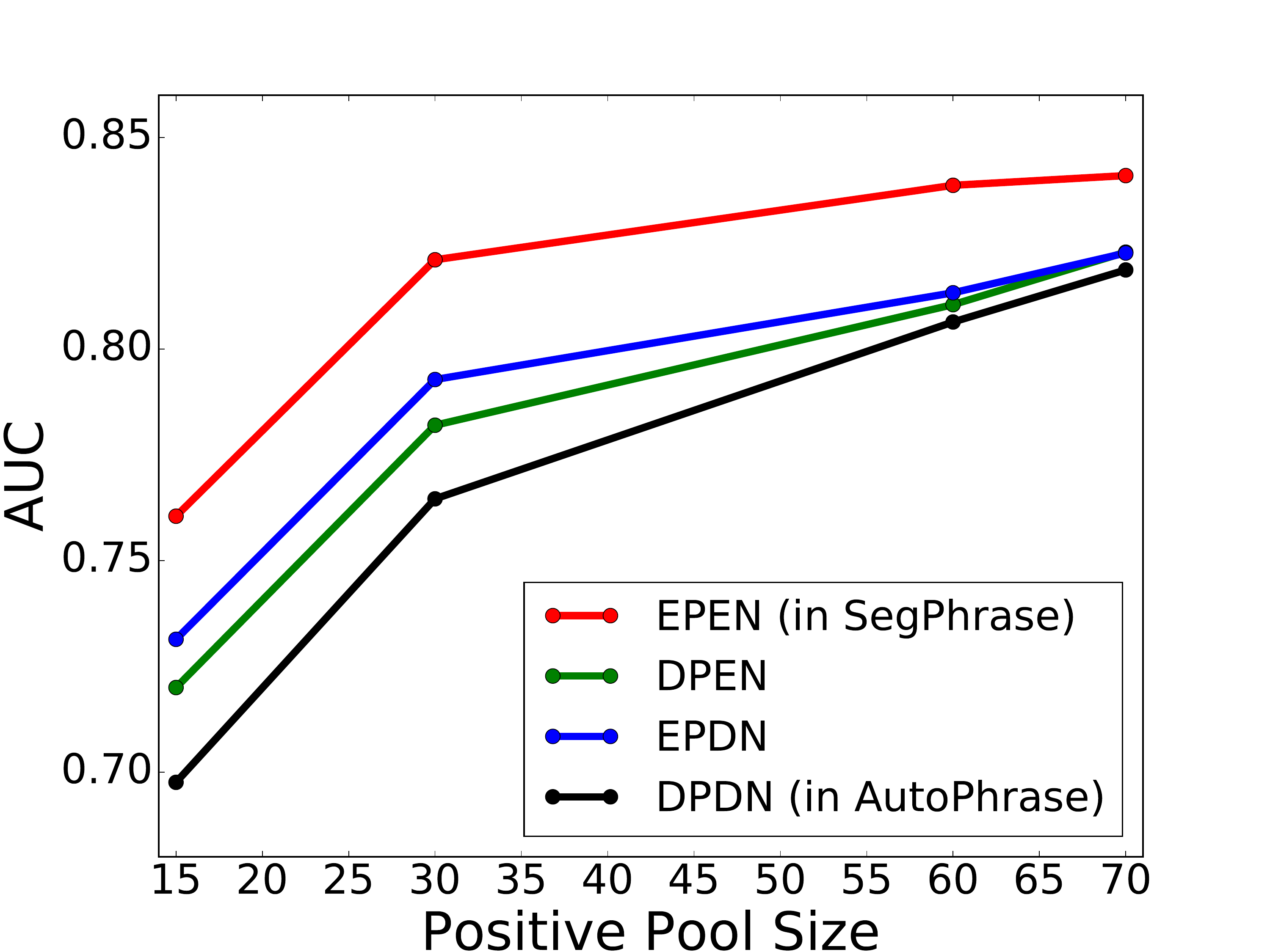}
      }
      \caption{AUC curves of four variants \emph{when we have enough positive labels in the positive pool \textbf{EP}}.}
      \label{fig:distant_1}
    \end{figure}

First of all, we evaluate the performance difference in the two positive pools.
Compared to \method{EPEN}, \method{DPEN} adopts a positive pool sampled from knowledge bases instead of the well-designed positive pool given by domain experts. The negative pool \emph{EN} is shared.
As shown in Figure~\ref{fig:distant_1}, we vary the size of the positive pool and plot their AUC curves.
We can find that \method{EPEN} outperforms \method{DPEN} and the trends of curves on both datasets are similar. Therefore, we conclude that the positive pool generated from knowledge bases has reasonable quality, although its corresponding quality estimator works slightly worse.

Secondly, we verify that whether the proposed noise reduction mechanism works properly.
Compared to \method{EPEN}, \method{EPDN} adopts a negative pool of all unlabeled phrase candidates instead of the well-designed negative pool given by domain experts. The positive pool \emph{EP} is shared.
In Figure~\ref{fig:distant_1}, the clear gap between them and the similar trends on both datasets show that the noisy negative pool is slightly worse than the well-designed negative pool, but it still works effectively.

As illustrated in Figure~\ref{fig:distant_1}, \method{DPDN} has the worst performance when the size of positive pools are limited. However, distant training can generate much larger positive pools, which may significantly beyond the ability of domain experts considering the high expense of labeling. Consequently, we are curious whether the distant training can finally beat domain experts when positive pool sizes become large enough. We call the size at this tipping point as the \term{ideal number}.

    \begin{figure}[t!]
      \centering
      \subfigure[DBLP]{
          \includegraphics[width= 0.22\textwidth]{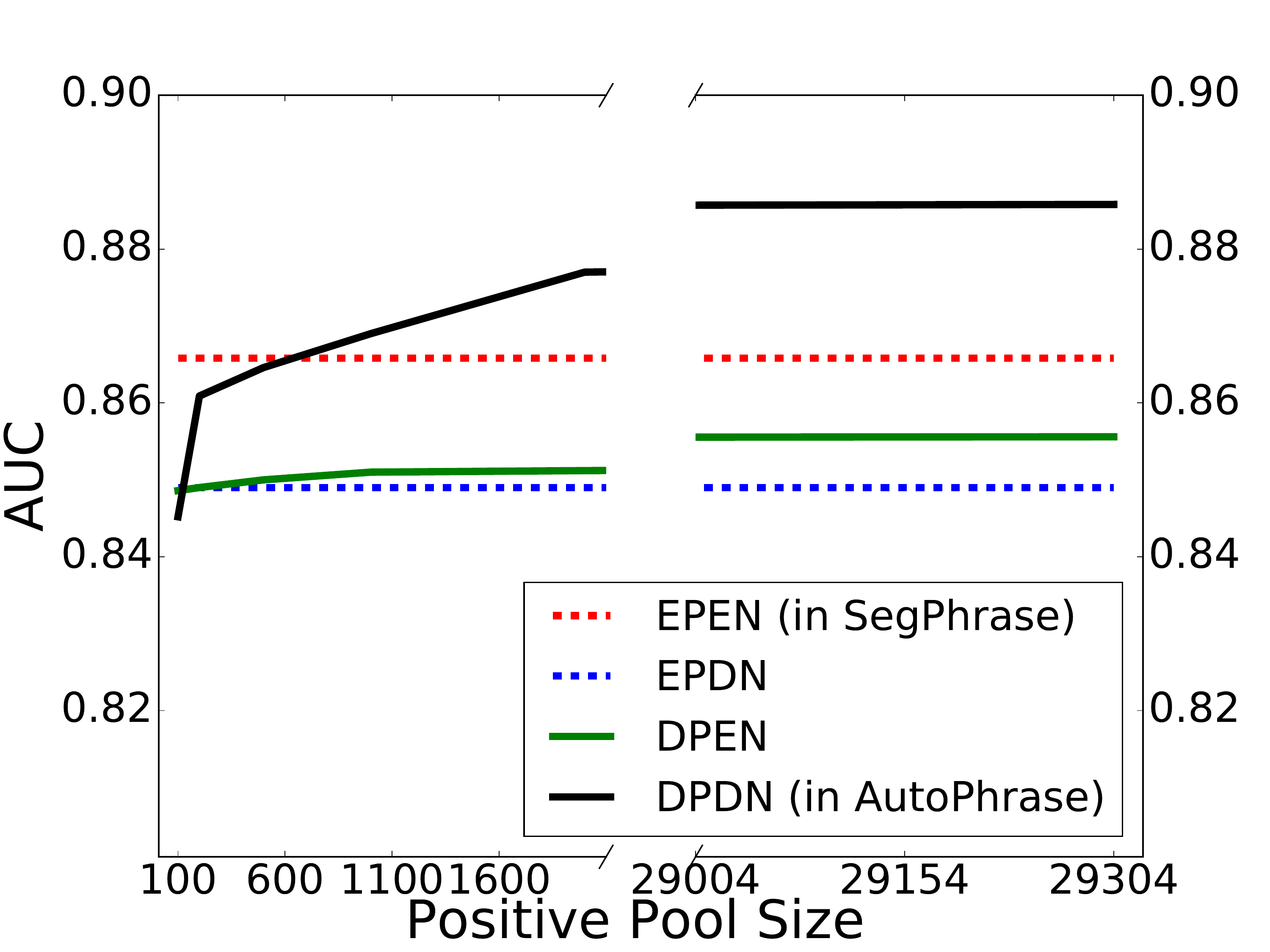}
      }
      \subfigure[Yelp]{
          \includegraphics[width= 0.22\textwidth]{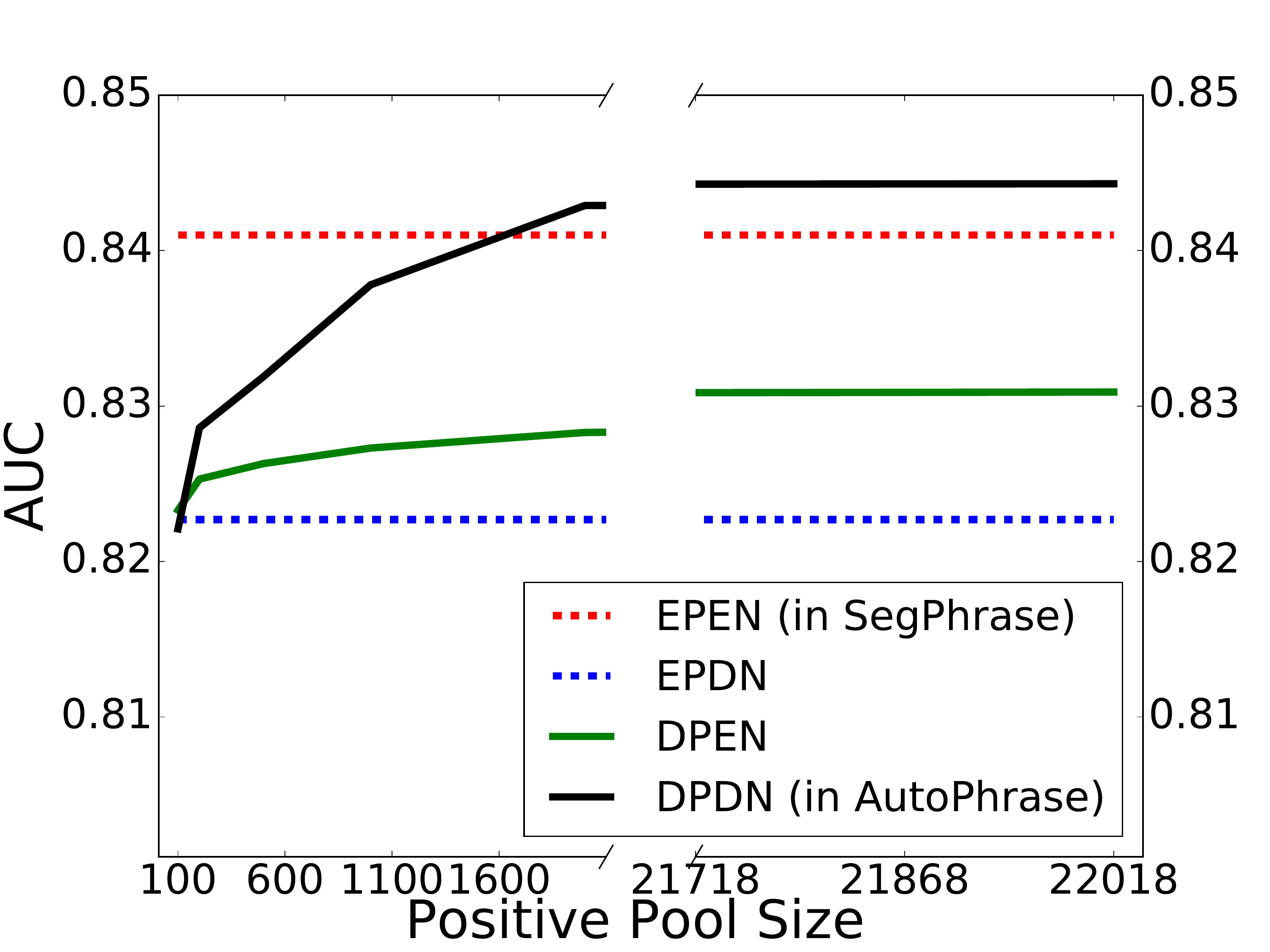}
      }
      \caption{AUC curves of four variants \emph{after we exhaust positive labels in the positive pool \textbf{EP}}.}
      \label{fig:distant_2}
    \end{figure}

We increase positive pool sizes and plot AUC curves of \method{DPEN} and \method{DPDN}, while \method{EPEN} and \method{EPDN} are degenerated as dashed lines due to the limited domain expert abilities. As shown in Figure~\ref{fig:distant_2}, with a large enough positive pool, distant training is able to beat expert labeling.
On the \emph{DBLP} dataset, the ideal number is about 700, while on the \emph{Yelp} dataset, it becomes around 1600. Our guess is that the ideal training size is proportional to the number of words (\eg, 91.6M in \emph{DBLP} and 145.1M in \emph{Yelp}). We notice that compared to the corpus size, the ideal number is relatively small, which implies the distant training should be effective in many domain-specific corpora as if they overlap with Wikipedia.

Besides, Figure~\ref{fig:distant_2} shows that when the positive pool size continues growing, the AUC score increases but the slope becomes smaller. The performance of distant training will be finally stable when a relatively large number of quality phrases were fed.

\begin{wrapfigure}{r}{0.23\textwidth}
  \centering
  \vspace{-0.3cm}
  \includegraphics[width=\linewidth]{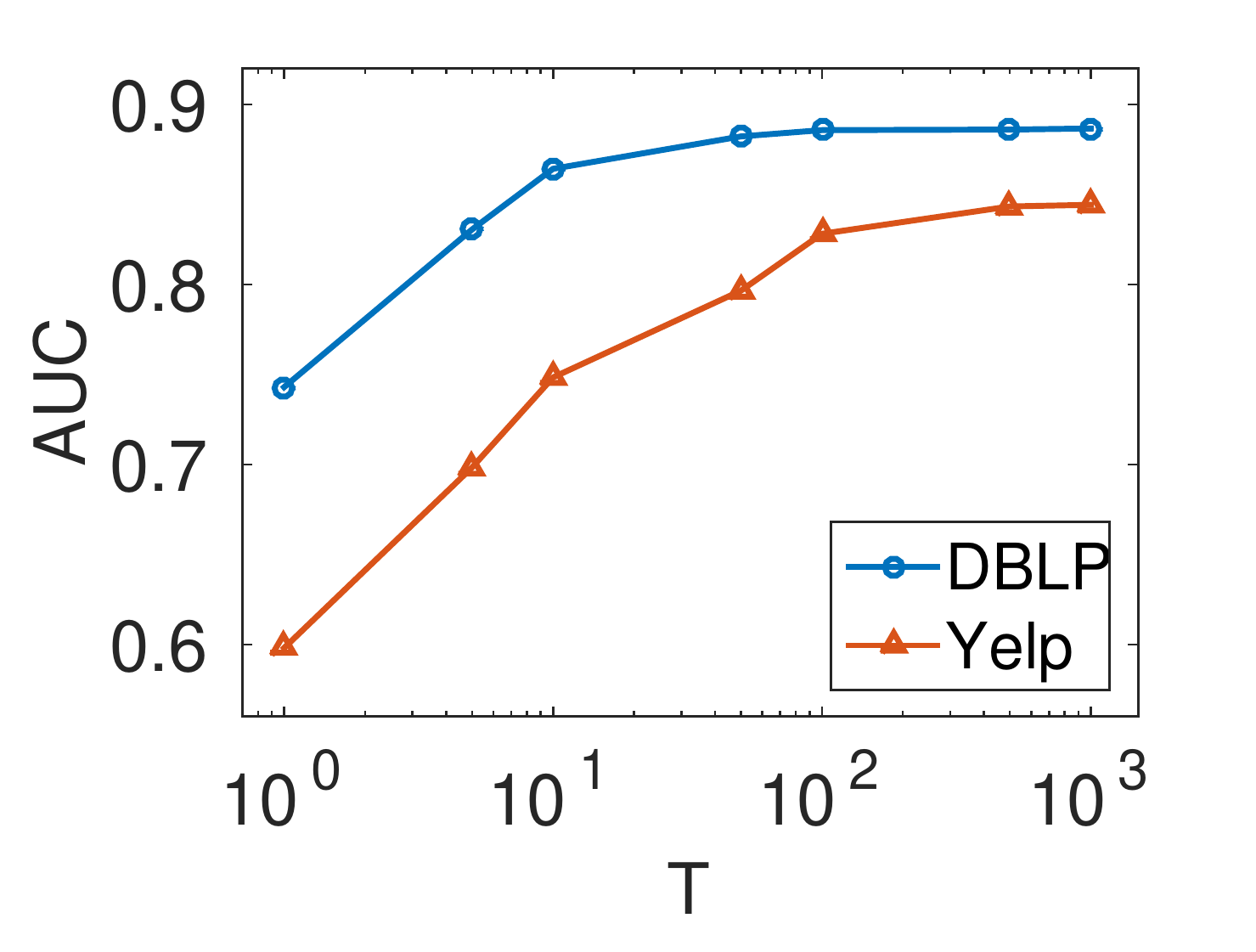}
  \vspace{-0.8cm}
  \caption{AUC curves of \method{DPDN} varying $T$.}\label{fig:distant_3}
  \vspace{-0.3cm}
\end{wrapfigure}
We are curious how many trees (\ie, $T$) is enough for \method{DPDN}.
We increase $T$ and plot AUC curves of \method{DPDN}.
As shown in Figure~\ref{fig:distant_3}, on both datasets, as $T$ grows, the AUC scores first increase rapidly and later the speed slows down gradually, which is consistent with the theoretical analysis in Section~\ref{sec:noisy_training}.

\subsection{POS-guided Phrasal Segmentation}

    We are also interested in how much performance gain we can obtain from incorporating POS tags in this segmentation model, especially for different languages. We select Wikipedia article datasets in three different languages: \emph{EN}, \emph{ES}, and \emph{CN}.

    \begin{figure}[t!]
      \centering
      \subfigure[EN]{
          \hspace{-3mm}\includegraphics[width= 0.16\textwidth]{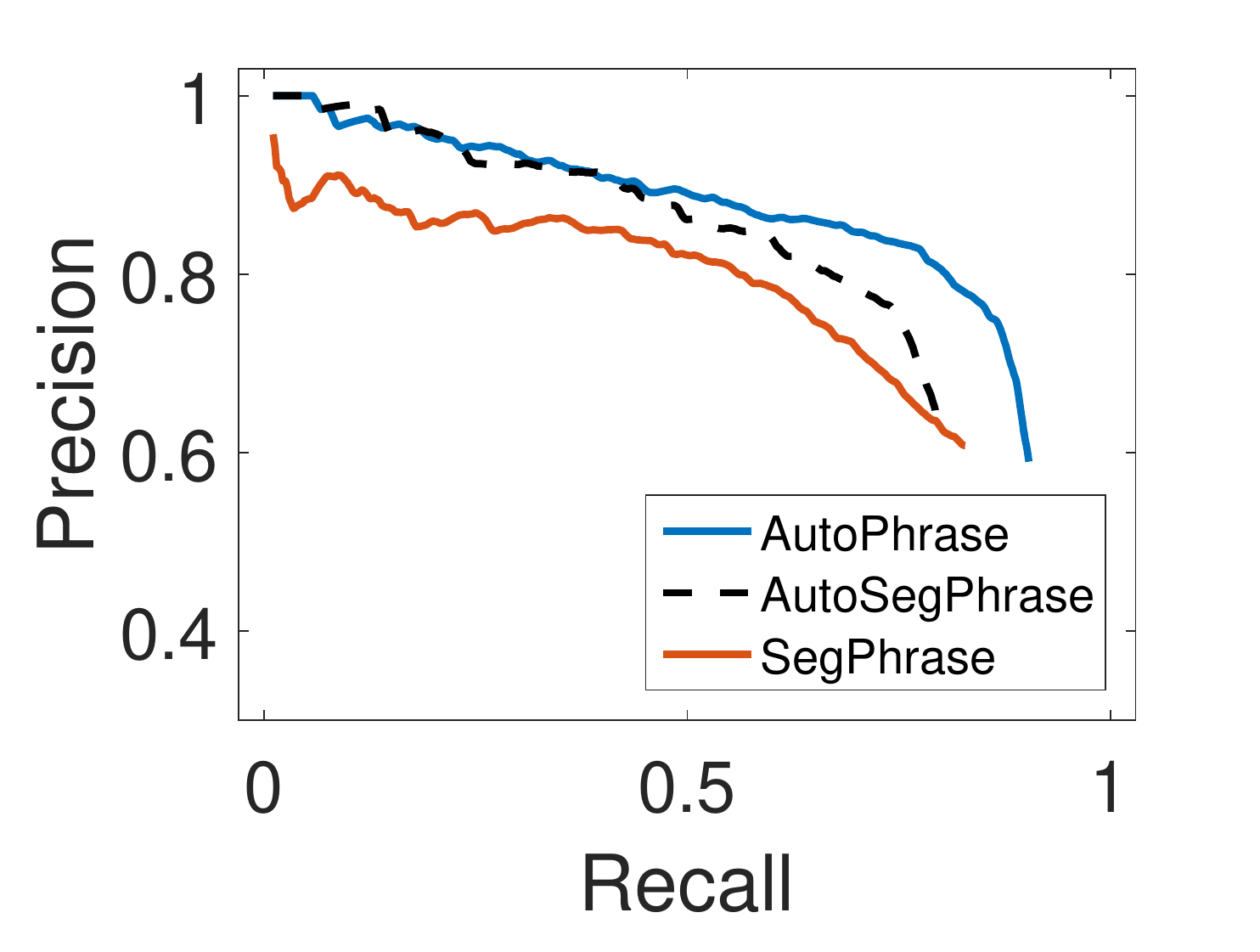}
      }
      \subfigure[ES]{
          \hspace{-4mm}\includegraphics[width= 0.16\textwidth]{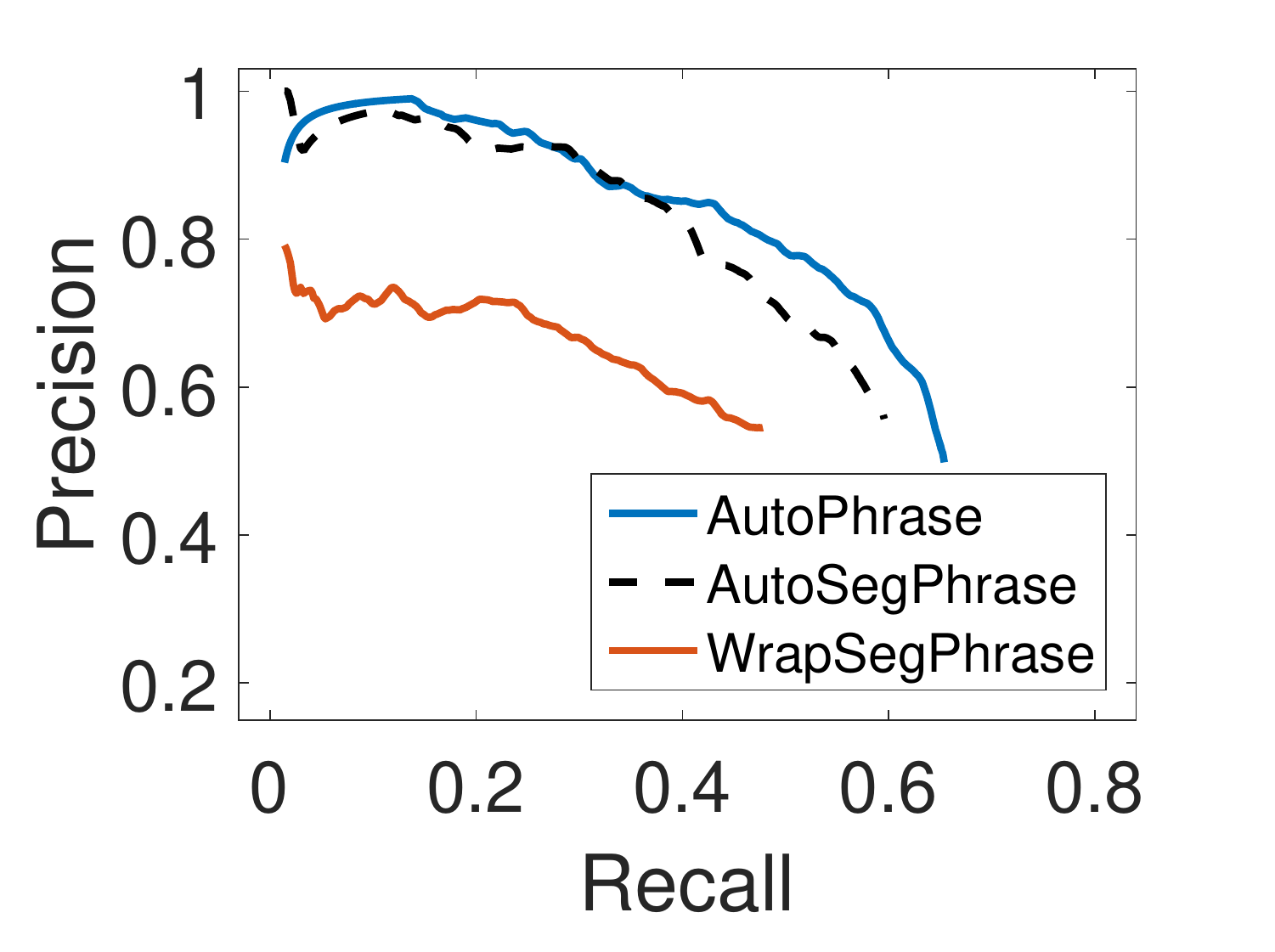}
      }
      \subfigure[CN]{
          \hspace{-4mm}\includegraphics[width= 0.16\textwidth]{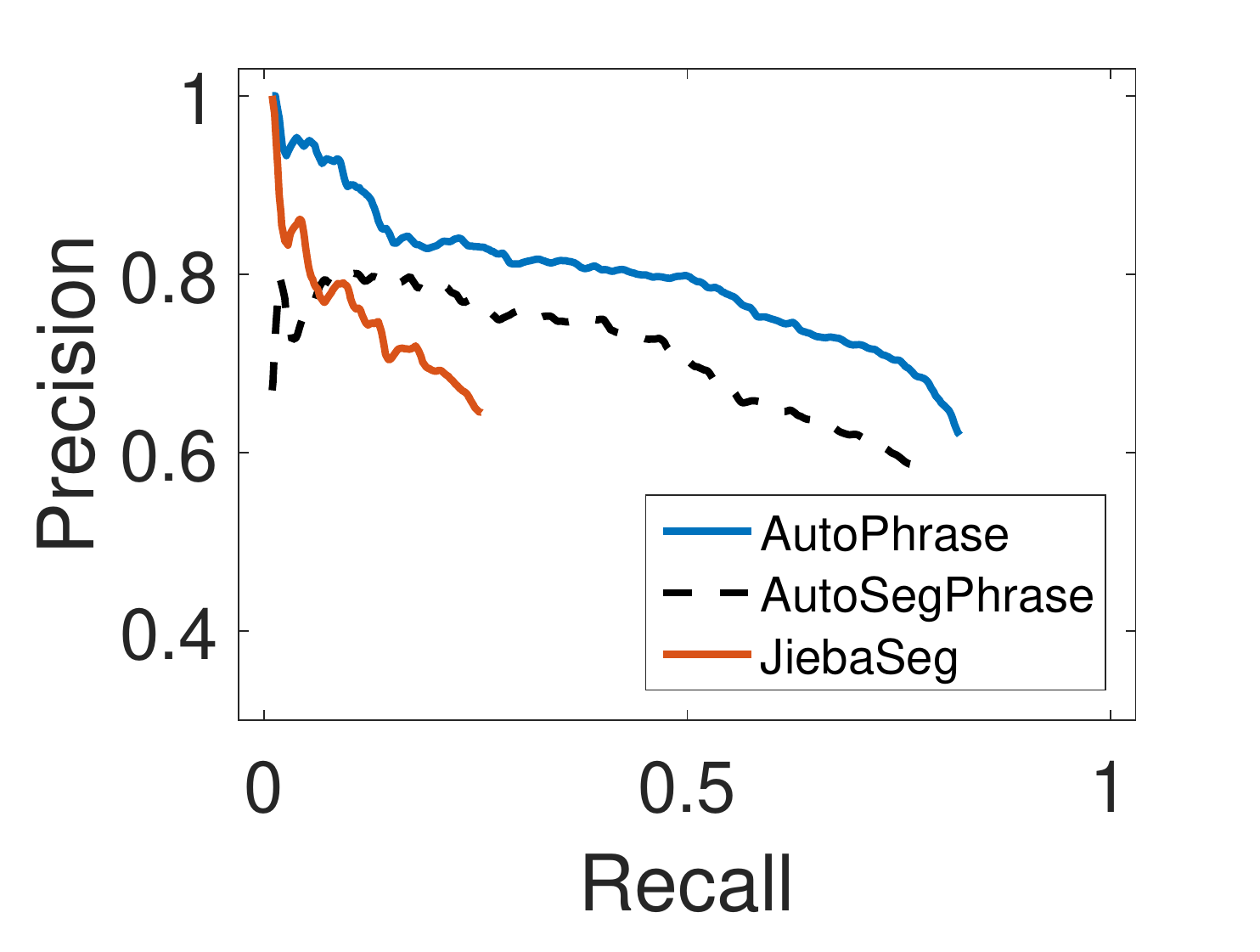}
      }
      \caption{Precision-recall curves of \LIPMine and \method{AutoSegPhrase}.}
      \label{fig:pos}
    \end{figure}

    Figure~\ref{fig:pos} compares the results of \LIPMine and \method{AutoSegPhrase}, with the best baseline methods as references.
    \LIPMine outperforms \method{AutoSegPhrase} even on the English dataset \emph{EN}, though it has been shown the length penalty works reasonably well in English~\cite{sigmod15_liu}. The Spanish dataset \emph{ES} has similar observation. Moreover, the advantage of \LIPMine becomes more significant on the \emph{CN} dataset, indicating the poor generality of length penalty.

    In summary, thanks to the extra context information and syntactic information for the particular language, incorporating POS tags during the phrasal segmentation can work better than equally penalizing phrases of the same length.


\begin{CJK}{UTF8}{gbsn}

\begin{table*}[t!]
\center
\caption{The results of \LIPMine on the \emph{EN} and \emph{CN} datasets, with translations and explanations for Chinese phrases. The whitespaces on the \emph{CN} dataset are inserted by the Chinese tokenizer.}
\label{tbl:case}
\begin{tabular}{|l|l|l|l|}
\hline
     & \multicolumn{1}{c|}{\emph{EN}} & \multicolumn{2}{c|}{\emph{CN}} \\
     \hline
Rank & Phrase & Phrase & Translation (Explanation) \\
\hline
1 & Elf Aquitaine & 江苏 舜 天 & (the name of a soccer team) \\
2 & Arnold Sommerfeld & 苦 艾 酒 & Absinthe \\
3 & Eugene Wigner & 白发 魔 女 & (the name of a novel/TV-series) \\
4 & Tarpon Springs & 笔记 型 电脑 & notebook computer, laptop \\
5 & Sean Astin &党委 书记 & Secretary of Party Committee \\
$\ldots$ & $\ldots$ & $\ldots$ & $\ldots$\\
20,001 & ECAC Hockey & 非洲 国家 & African countries \\
20,002 & Sacramento Bee & 左翼 党 & The Left (German: Die Linke)\\
20,003 & Bering Strait & 菲 沙 河谷 & Fraser Valley\\
20,004 & Jacknife Lee & 海马 体 & Hippocampus\\
20,005 & WXYZ-TV & 斋 贺光希 & Mitsuki Saiga (a voice actress)\\
$\ldots$ & $\ldots$ & $\ldots$ & $\ldots$\\
99,994 & John Gregson & 计算机 科学技术 & Computer Science and Technology\\
99,995 & white-tailed eagle & 恒 天然 & Fonterra (a company) \\
99,996 & rhombic dodecahedron & 中国 作家 协会 & The Vice President of Writers \\
       &                      &   副 主席      & Association of China\\
99,997 & great spotted woodpecker & 维他命 b & Vitamin B\\
99,998 & David Manners & 舆论 导向 & controlled guidance of the media\\
$\ldots$ & $\ldots$ & $\ldots$ & $\ldots$\\
\hline
\end{tabular}
\end{table*}
\end{CJK}

\subsection{Case Study}\label{sec:case}

\begin{CJK}{UTF8}{gbsn}

We present a case study about the extracted phrases as shown in Table~\ref{tbl:case}. The top ranked phrases are mostly named entities, which makes sense for the Wikipedia article datasets. Even in the long tail part, there are still many high-quality phrases. For example, we have the $\lceil$great spotted woodpecker$\rfloor$ (a type of birds) and $\lceil$计算机 科学技术$\rfloor$ (\ie, Computer Science and Technology) ranked about 100,000. In fact, we have more than 345K and 116K phrases with a phrase quality higher than $0.5$ on the \emph{EN} and \emph{CN} datasets respectively.

\end{CJK}

\subsection{Efficiency Evaluation}

    To study both time and memory efficiency, we choose the three largest datasets: \emph{EN}, \emph{ES}, and \emph{CN}.

    \begin{figure}[t]
      \centering
      \subfigure[Running Time]{
        \label{fig:time}
        \includegraphics[width = 0.14\textwidth]{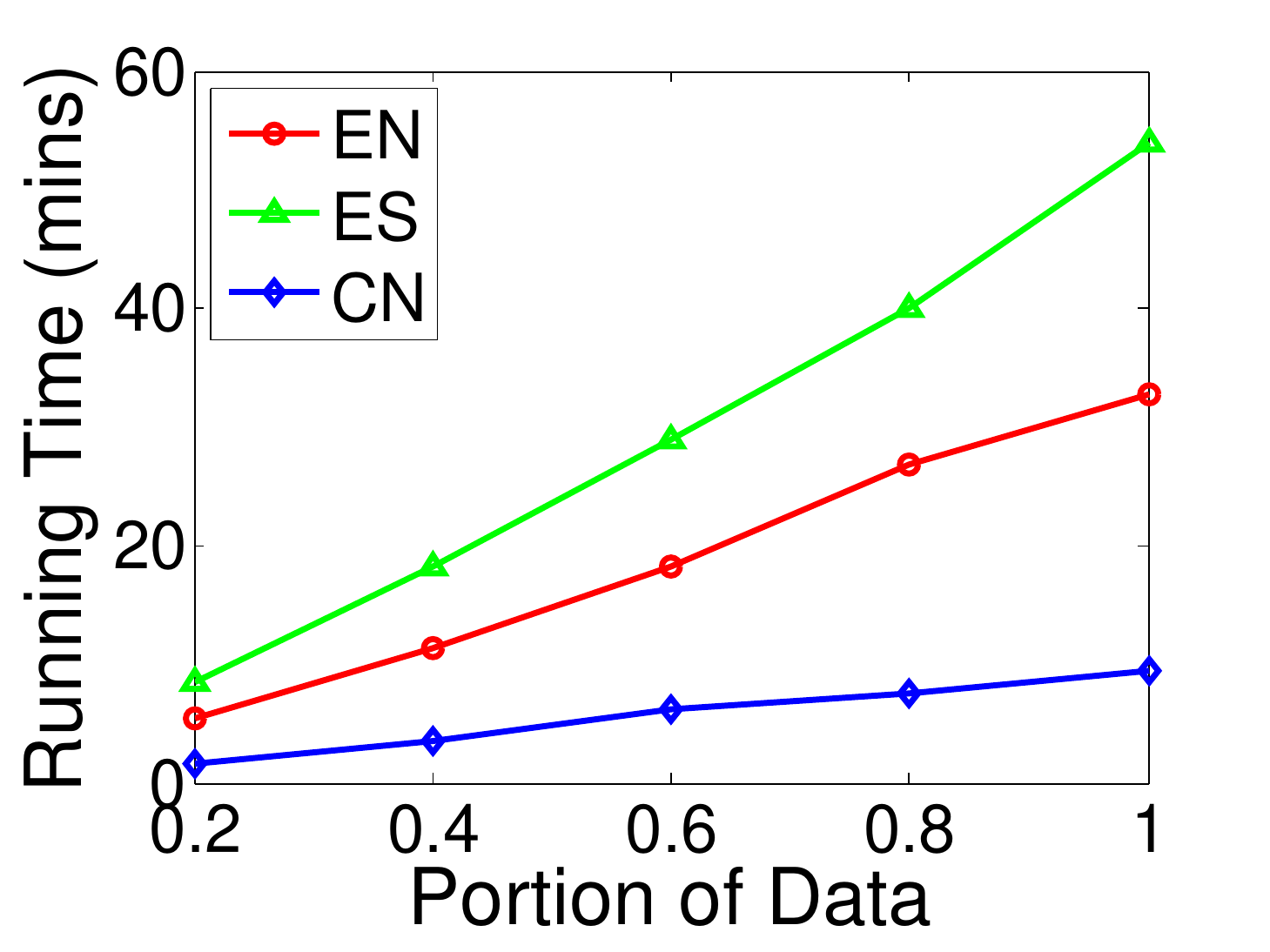}
      }
      \subfigure[Peak Memory]{
        \label{fig:mem}
        \includegraphics[width = 0.14\textwidth]{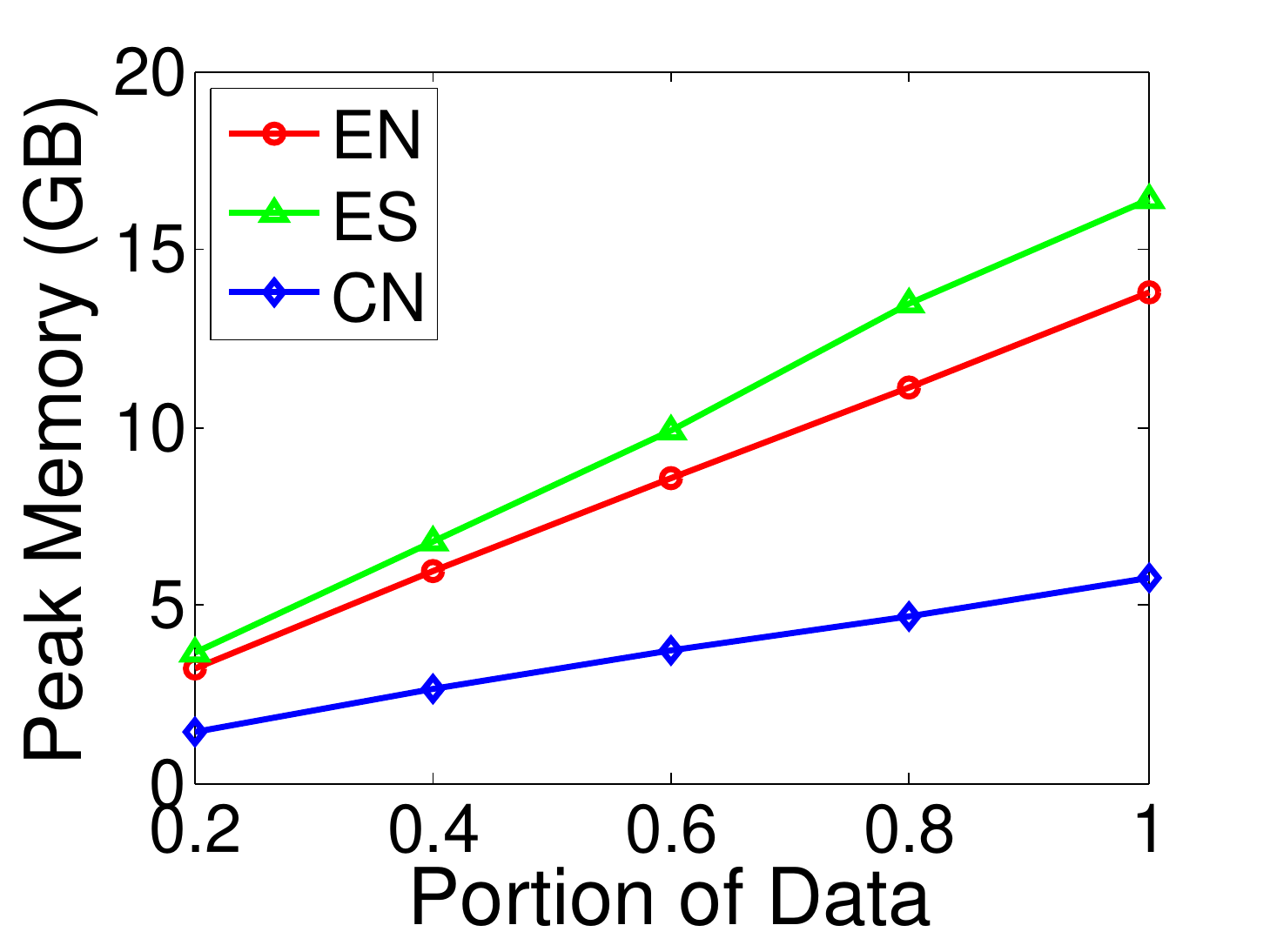}
      }
      \subfigure[Multi-threading]{
        \label{fig:speedup}
        \includegraphics[width = 0.14\textwidth]{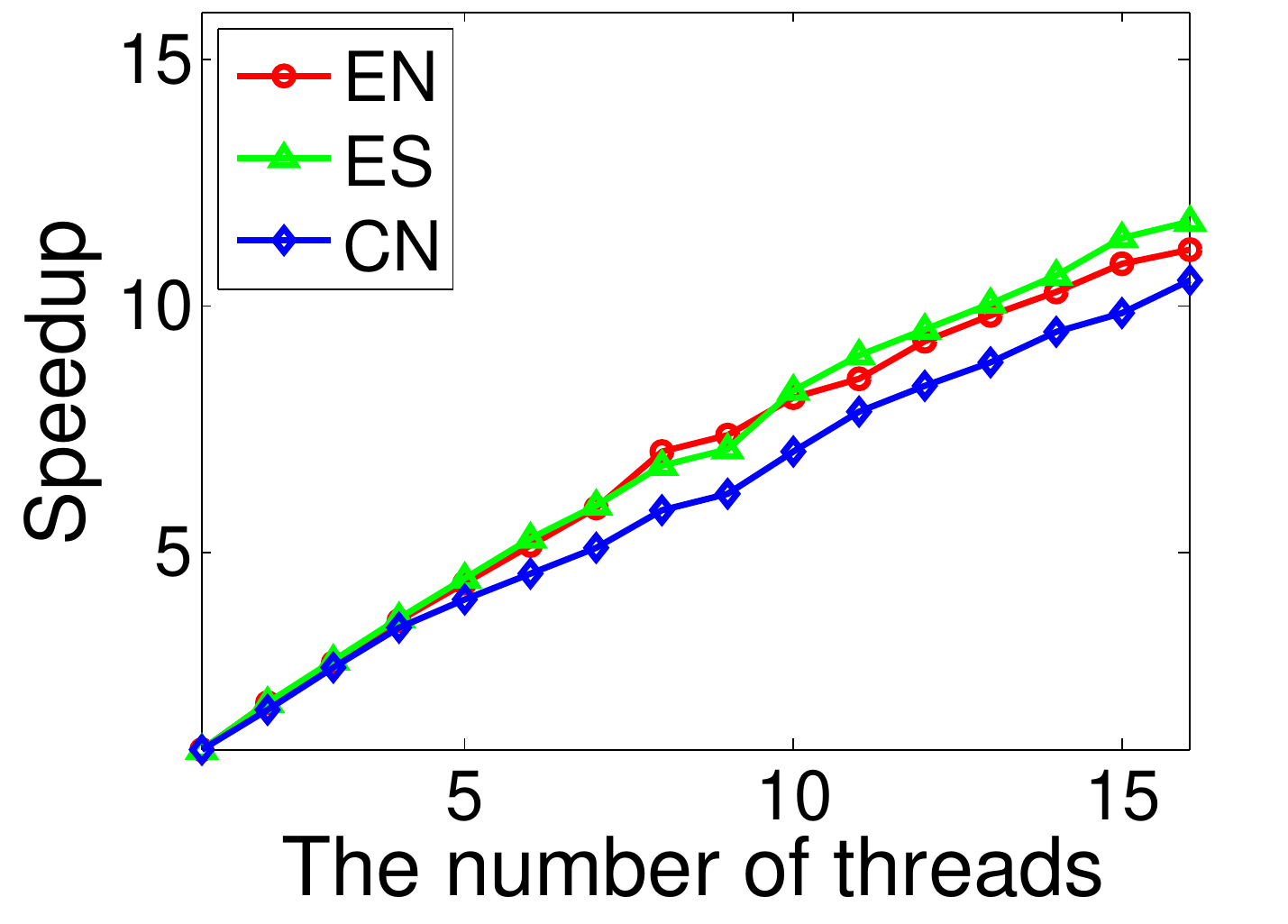}
      }
      \caption{Efficiency of \LIPMine.}
    \end{figure}

    Figures~\ref{fig:time} and~\ref{fig:mem} evaluate the running time and the peak memory usage of \LIPMine using 10 threads on different proportions of three datasets respectively. Both time and memory are linear to the size of text corpora. Moreover, \LIPMine can also be parallelized in an almost lock-free way and shows a linear speedup in Figure~\ref{fig:speedup}.

    \begin{table*}[t]
        \center
        \caption{Efficiency Comparison between \LIPMine and \SegPhrase/\method{WrapSegPhrase} utilizing 10 threads.}
        \label{tbl:segphrase}
        \begin{tabular}{|l|c|c|c|c|c|c|}
        \hline
            & \multicolumn{2}{c|}{EN} & \multicolumn{2}{c|}{ES} & \multicolumn{2}{c|}{CN} \\
        \hline
            & Time & Memory & Time & Memory & Time & Memory \\
            & (mins) & (GB) & (mins) & (GB) & (mins) & (GB) \\
        \hline
        \LIPMine & 32.77 & 13.77 & 54.05 & 16.42 & 9.43 & 5.74 \\
        \hline
        \method{(Wrap)SegPhrase} & 369.53 & 97.72 & 452.85 & 92.47 & 108.58 & 35.38\\
        \hline
        Speedup/Saving & 11.27 & 86\% & 8.37 & 82\% & 11.50 & 83\% \\
        \hline
        \end{tabular}
    \end{table*}

    Besides, compared to the previous state-of-the-art phrase mining method \SegPhrase and its variants \method{WrapSegPhrase} on three datasets, as shown in Table~\ref{tbl:segphrase}, \LIPMine achieves about 8 to 11 times speedup and about 5 to 7 times memory usage improvement. These improvements are made by a more efficient indexing and a more thorough parallelization.

    \begin{figure*}[t!]
      \centering
      \subfigure[EN]{
        \includegraphics[width = 0.3\textwidth]{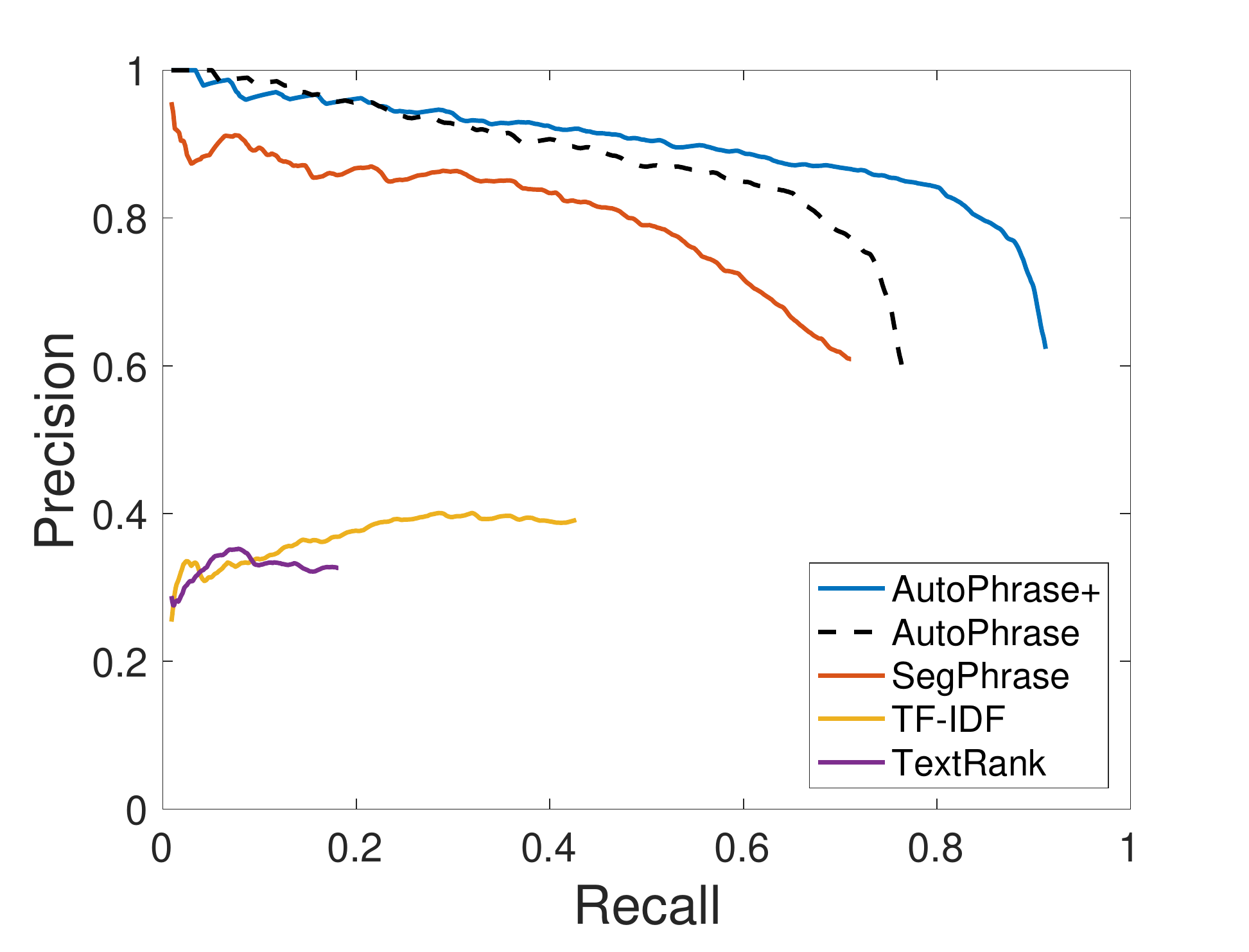}
      }
      \subfigure[ES]{
        \includegraphics[width = 0.3\textwidth]{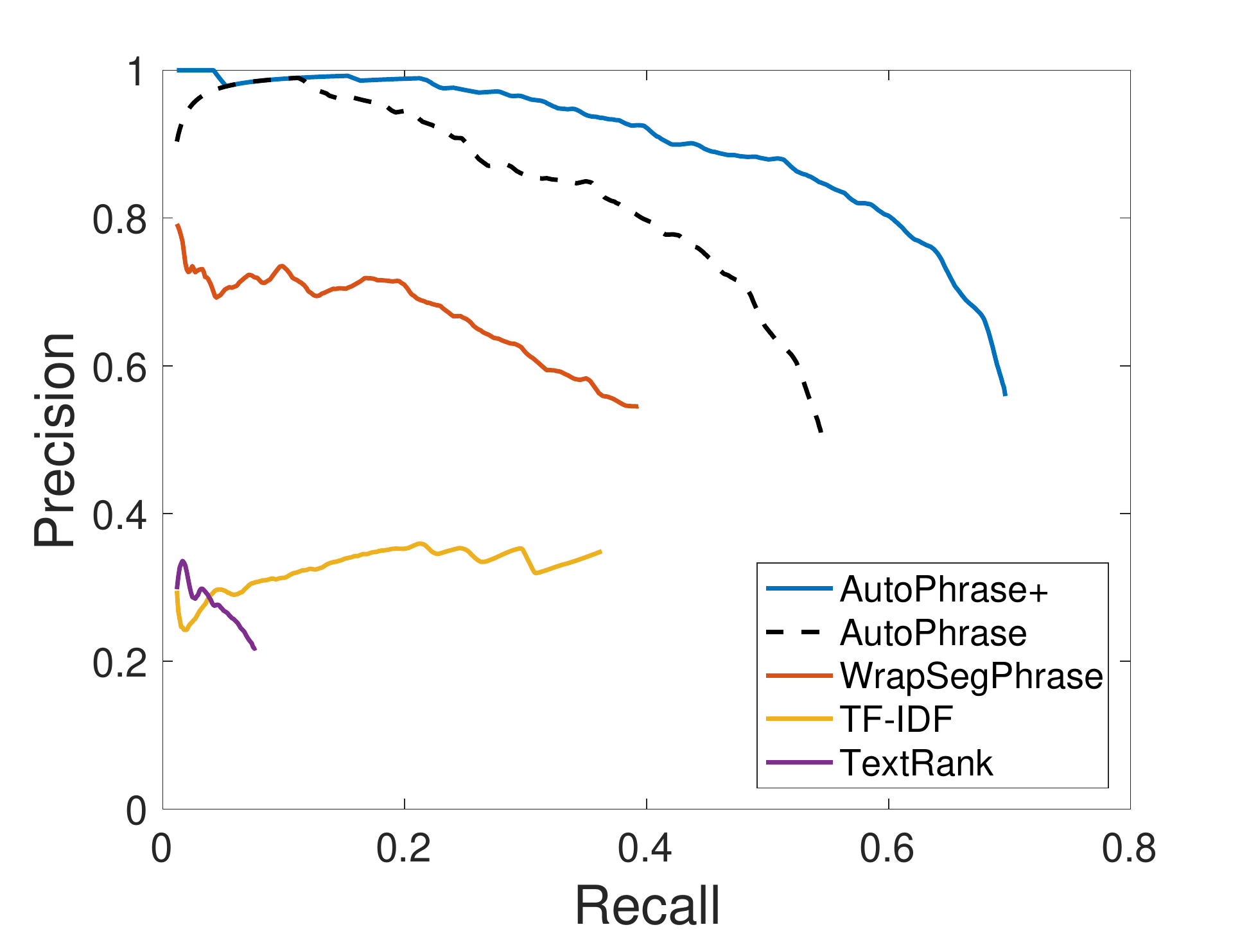}
      }
      \subfigure[CN]{
        \label{fig:speedup}
        \includegraphics[width = 0.3\textwidth]{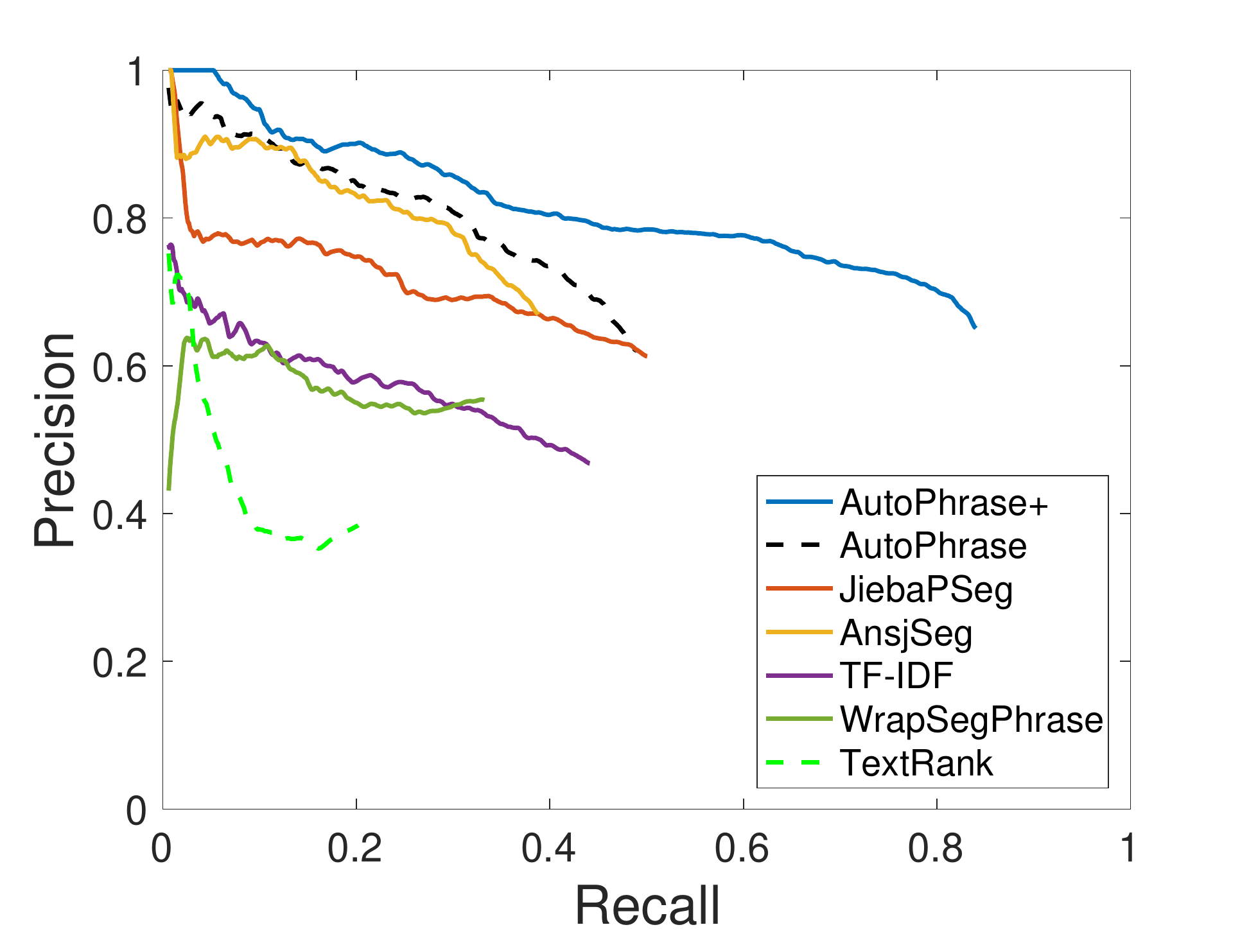}
      }
      \caption{Precision-recall curves evaluated by human annotation with both single-word and multi-word phrases in pools.}
      \label{fig:single}
    \end{figure*}

\section{Single-Word Phrases}

\label{sec:single}

\LIPMine can be extended to model single-word phrases, which can gain about $10\%$ to $30\%$ recall improvements on different datasets. To study the effect of modeling quality single-word phrases, we choose the three Wikipedia article datasets in different languages: \emph{EN}, \emph{ES}, and \emph{CN}.

    \subsection{Quality Estimation}
    In \emph{the paper}, the definition of quality phrases and the evaluation only focus on multi-word phrases. In linguistic analysis, however, a phrase is not only a group of multiple words, but also possibly a single word, as long as it functions as a constituent in the syntax of a sentence~\cite{finch2000linguistic}. As a great portion (ranging from $10\%$ to $30\%$ on different datasets based on our experiments) of high-quality phrases, we should take single-word phrases (\eg, \phrase{UIUC}, \phrase{Illinois}, and \phrase{USA}) into consideration as well as multi-word phrases to achieve a high recall in phrase mining.

    Considering the criteria of quality phrases, because single-word phrases cannot be decomposed into two or more parts, the \emph{concordance} and \emph{completeness} are no longer definable. Therefore, we revise the requirements for \term{quality single-word phrases} as below.
    \begin{itemize}[leftmargin=*,noitemsep,nolistsep]
    \item \term{Popularity}: Quality phrases should occur with sufficient frequency in the given document collection.
    \item \term{Informativeness}: A phrase is informative if it is indicative of a specific topic or concept.
    \item \term{Independence}: A quality single-word phrase is more likely a complete semantic unit in the given documents.
    \end{itemize}
    Only single-word phrases satisfying all \emph{popularity}, \emph{independence}, and \emph{informativeness} requirements are recognized as quality single-word phrases.
    \begin{thm:eg}
    ``\emph{UIUC}'' is a quality single-word phrase.
    ``\emph{this}'' is not a quality phrase due to its low informativeness.
    ``\emph{united}'', usually occurring within other quality multi-word phrases such as ``\emph{United States}'', ``\emph{United Kingdom}'', ``\emph{United Airlines}'', and ``\emph{United Parcel Service}'', is not a quality single-word phrase, because its independence is not enough.
    \end{thm:eg}

    After the phrasal segmentation, in replacement of concordance features, the \textbf{independence feature} is added for single-word phrases. Formally, it is the ratio of the rectified frequency of a single-word phrase given the phrasal segmentation over its raw frequency. Quality single-word phrases are expected to have large values. For example, ``\emph{united}'' is likely to an almost zero ratio.

    We use \textbf{AutoPhrase+} to denote the extended \LIPMine with quality single-word phrase estimation.

    \subsection{Experiments}

    We have a similar human annotation as that in \emph{the paper}. Differently, we randomly sampled $500$ Wiki-uncovered phrases from the returned phrases (\emph{both single-word and multi-word phrases}) of each method in experiments of \emph{the paper}. Therefore, we have \emph{new pools} on the \emph{EN}, \emph{ES}, and \emph{CN} datasets.
    The intra-class correlations (ICCs) are all more than 0.9, which shows the agreement.

    Figure~\ref{fig:single} compare all methods based these new pools. Note that all methods except for \SegPhrase/\method{WrapSegPhrase} extract single-word phrases as well.

    Significant recall advantages can be always observed on all \emph{EN}, \emph{ES}, and \emph{CN} datasets. The recall differences between \method{AutoPhrase+} and \LIPMine, ranging from 10\% to 30\% sheds light on the importance of modeling single-word phrases.
    Across two Latin language datasets, \emph{EN} and \emph{ES}, \method{AutoPhrase+} and \LIPMine overlaps in the beginning, but later, the precision of \LIPMine drops earlier and has a lower recall due to the lack of single-word phrases. On the \emph{CN} dataset, \method{AutoPhrase+} and \LIPMine has a clear gap even in the very beginning, which is different from the trends on the \emph{EN} and \emph{ES} datasets, which reflects that single-word phrases are more important in Chinese. The major reason behind is that there are a considerable number of high-quality phrases (\eg, person names) in Chinese have only one token after tokenization.


\section{Conclusions}\label{sec:con}

In this paper, we present an automated phrase mining framework with two novel techniques: the robust positive-only distant training and the POS-guided phrasal segmentation incorporating part-of-speech (POS) tags, for the development of an \emph{automated phrase mining} framework \LIPMine.
Our extensive experiments show that \LIPMine is domain-independent, outperforms other phrase mining methods, and supports multiple languages (\eg, English, Spanish, and Chinese) effectively, with \MHE.

Besides, the inclusion of quality single-word phrases (\eg, $\lceil$UIUC$\rfloor$ and $\lceil$USA$\rfloor$) which leads to about 10\% to 30\% increased recall and the exploration of better indexing strategies and more thorough parallelization, which leads to about 8 to 11 times running time speedup and about 80\% to 86\% memory usage saving over \SegPhrase.
Interested readers may try our released code at GitHub.

For future work, it is interesting to (1) refine quality phrases to entity mentions, (2) apply \LIPMine to more languages, such as Japanese, and (3) for those languages without general knowledge bases, seek an unsupervised method to generate the positive pool from the corpus, even with some noise.

\bibliographystyle{abbrv}
\bibliography{main}

\begin{thebibliography}{10}

\bibitem{allahverdyan2011comparative}
A.~Allahverdyan and A.~Galstyan.
\newblock Comparative analysis of viterbi training and maximum likelihood
  estimation for hmms.
\newblock In {\em NIPS}, pages 1674--1682, 2011.

\bibitem{breiman2000randomizing}
L.~Breiman.
\newblock Randomizing outputs to increase prediction accuracy.
\newblock {\em Machine Learning}, 40(3):229--242, 2000.

\bibitem{chen1994extracting}
K.-h. Chen and H.-H. Chen.
\newblock Extracting noun phrases from large-scale texts: A hybrid approach and
  its automatic evaluation.
\newblock In {\em ACL}, 1994.

\bibitem{de2006generating}
M.-C. De~Marneffe, B.~MacCartney, C.~D. Manning, et~al.
\newblock Generating typed dependency parses from phrase structure parses.
\newblock In {\em Proceedings of LREC}, volume~6, pages 449--454, 2006.

\bibitem{deane2005nonparametric}
P.~Deane.
\newblock A nonparametric method for extraction of candidate phrasal terms.
\newblock In {\em ACL}, 2005.

\bibitem{ahmedTopMine2015}
A.~El-Kishky, Y.~Song, C.~Wang, C.~R. Voss, and J.~Han.
\newblock Scalable topical phrase mining from text corpora.
\newblock {\em VLDB}, 8(3), Aug. 2015.

\bibitem{evans1996noun}
D.~A. Evans and C.~Zhai.
\newblock Noun-phrase analysis in unrestricted text for information retrieval.
\newblock In {\em Proceedings of the 34th annual meeting on Association for
  Computational Linguistics}, pages 17--24. Association for Computational
  Linguistics, 1996.

\bibitem{finch2000linguistic}
G.~Finch.
\newblock {\em Linguistic terms and concepts}.
\newblock Macmillan Press Limited, 2000.

\bibitem{frantzi2000automatic}
K.~Frantzi, S.~Ananiadou, and H.~Mima.
\newblock Automatic recognition of multi-word terms:. the c-value/nc-value
  method.
\newblock {\em JODL}, 3(2):115--130, 2000.

\bibitem{geurts2006extremely}
P.~Geurts, D.~Ernst, and L.~Wehenkel.
\newblock Extremely randomized trees.
\newblock {\em Machine learning}, 63(1):3--42, 2006.

\bibitem{koo2008simple}
T.~Koo, X.~Carreras, and M.~Collins.
\newblock Simple semi-supervised dependency parsing.
\newblock {\em ACL-HLT}, 2008.

\bibitem{levy2003harder}
R.~Levy and C.~Manning.
\newblock Is it harder to parse chinese, or the chinese treebank?
\newblock In {\em Proceedings of the 41st Annual Meeting on Association for
  Computational Linguistics-Volume 1}, pages 439--446. Association for
  Computational Linguistics, 2003.

\bibitem{sigmod15_liu}
J.~Liu, J.~Shang, C.~Wang, X.~Ren, and J.~Han.
\newblock Mining quality phrases from massive text corpora.
\newblock In {\em Proceedings of 2015 ACM SIGMOD International Conference on
  Management of Data}, 2015.

\bibitem{liu2011automatic}
Z.~Liu, X.~Chen, Y.~Zheng, and M.~Sun.
\newblock Automatic keyphrase extraction by bridging vocabulary gap.
\newblock In {\em Proceedings of the Fifteenth Conference on Computational
  Natural Language Learning}, pages 135--144. Association for Computational
  Linguistics, 2011.

\bibitem{martinez2005switching}
G.~Mart{\'\i}nez-Mu{\~n}oz and A.~Su{\'a}rez.
\newblock Switching class labels to generate classification ensembles.
\newblock {\em Pattern Recognition}, 38(10):1483--1494, 2005.

\bibitem{mcdonald2005non}
R.~McDonald, F.~Pereira, K.~Ribarov, and J.~Haji{\v{c}}.
\newblock Non-projective dependency parsing using spanning tree algorithms.
\newblock In {\em EMNLP}, 2005.

\bibitem{mihalcea2004textrank}
R.~Mihalcea and P.~Tarau.
\newblock Textrank: Bringing order into texts.
\newblock In {\em ACL}, 2004.

\bibitem{nivre2016universal}
J.~Nivre, M.-C. de~Marneffe, F.~Ginter, Y.~Goldberg, J.~Hajic, C.~D. Manning,
  R.~McDonald, S.~Petrov, S.~Pyysalo, N.~Silveira, et~al.
\newblock Universal dependencies v1: A multilingual treebank collection.
\newblock In {\em Proceedings of the 10th International Conference on Language
  Resources and Evaluation (LREC 2016)}, 2016.

\bibitem{Aditya10}
A.~Parameswaran, H.~Garcia-Molina, and A.~Rajaraman.
\newblock Towards the web of concepts: Extracting concepts from large datasets.
\newblock {\em Proceedings of the Very Large Data Bases Conference (VLDB)},
  3((1-2)), September 2010.

\bibitem{park2002automatic}
Y.~Park, R.~J. Byrd, and B.~K. Boguraev.
\newblock Automatic glossary extraction: beyond terminology identification.
\newblock In {\em COLING}, 2002.

\bibitem{punyakanok2001use}
V.~Punyakanok and D.~Roth.
\newblock The use of classifiers in sequential inference.
\newblock In {\em NIPS}, 2001.

\bibitem{schmid1995treetagger}
H.~Schmid.
\newblock Treetagger| a language independent part-of-speech tagger.
\newblock {\em Institut f{\"u}r Maschinelle Sprachverarbeitung, Universit{\"a}t
  Stuttgart}, 43:28, 1995.

\bibitem{witten1999kea}
I.~H. Witten, G.~W. Paynter, E.~Frank, C.~Gutwin, and C.~G. Nevill-Manning.
\newblock Kea: Practical automatic keyphrase extraction.
\newblock In {\em Proceedings of the fourth ACM conference on Digital
  libraries}, pages 254--255. ACM, 1999.

\bibitem{xun2000unified}
E.~Xun, C.~Huang, and M.~Zhou.
\newblock A unified statistical model for the identification of english basenp.
\newblock In {\em ACL}, 2000.

\bibitem{zhang2008comparative}
Z.~Zhang, J.~Iria, C.~A. Brewster, and F.~Ciravegna.
\newblock A comparative evaluation of term recognition algorithms.
\newblock {\em LREC}, 2008.

\end{thebibliography}

\end{document}